Published at the 21st Machine Learning in Computational Biology (MLCB 2026)

# EvoFlows: Evolutionary Edit-Based Flow-Matching for Protein Engineering

**Nicolas Deutschmann*** , **Constance Ferragu***, **Jonathan D. Ziegler***, **Shayan Aziznejad & Eli Bixby**
Cradle
Zürich, CH
{nicolas, constance, jonathan}@cradle.bio

## Abstract

Protein optimization improves an existing protein through targeted mutations, but existing protein language models are poorly suited to it: autoregressive models require full sequence generation, masked language and discrete diffusion models rely on pre-specified mutation locations, and none naturally support insertions and deletions relative to a template. We introduce EvoFlows, a variable-length sequence-to-sequence approach that learns mutational trajectories between evolutionarily related proteins via edit flows, predicting not only *which* mutation to perform, but also *where* and *how many* should occur. Across diverse protein families from UniRef and OAS, EvoFlows generates variants that remain consistent with natural families while exploring farther from the template than leading baselines. *In vitro* validation of a VHH binder and an enzyme confirms that the large majority of generated variants remain functional and stable without computational downselection.

## 1 Introduction

Protein optimization aims to improve the properties of an existing protein through targeted mutations (substitutions, insertions, and deletions) to its sequence. Unlike *de-novo* design, which seeks to generate entirely new sequences, optimization starts from a known functional protein that already satisfies certain requirements, such as binding to a target or exhibiting measurable activity. The goal is then to explore its local neighborhood by applying mutations that preserve overall structure and function while improving targeted properties (Listov et al., 2024), a key step in pre-clinical drug development pipelines (Paul et al., 2010).

In recent years, protein language models (PLMs) have emerged as successful tools for protein optimization (Hie et al., 2024; Lin et al., 2023; Madani et al., 2023). While PLMs capture rich functional and evolutionary information, existing models are not naturally aligned with the local-mutation setting under which optimization operates. Autoregressive models regenerate an entire protein one residue at a time, making it difficult to control how many mutations are introduced and therefore how far the generated sequence drifts from the starting protein. This is particularly limiting in optimization settings, where the goal is often to introduce only a few local mutations. Masked language models are better suited for local editing, but require substitution sites to be specified externally and do not support insertions, deletions, or variable-length generation.

Protein optimization would therefore benefit from models that combine the local-editing capabilities of masked language models with the variable-length generation abilities of autoregressive models, without requiring externally specified mutation locations. Native support for insertions and deletions is particularly important for protein families such as antibodies, where indels are common (Rock et al., 1994). Before being conditioned on a specific task (Gruver et al., 2023; Widatalla et al., 2024), such a model should first generate function-preserving variants of a protein for unguided exploration (Hie et al., 2024) and serve as a strong foundation for fine-tuning.

*Equal contribution.

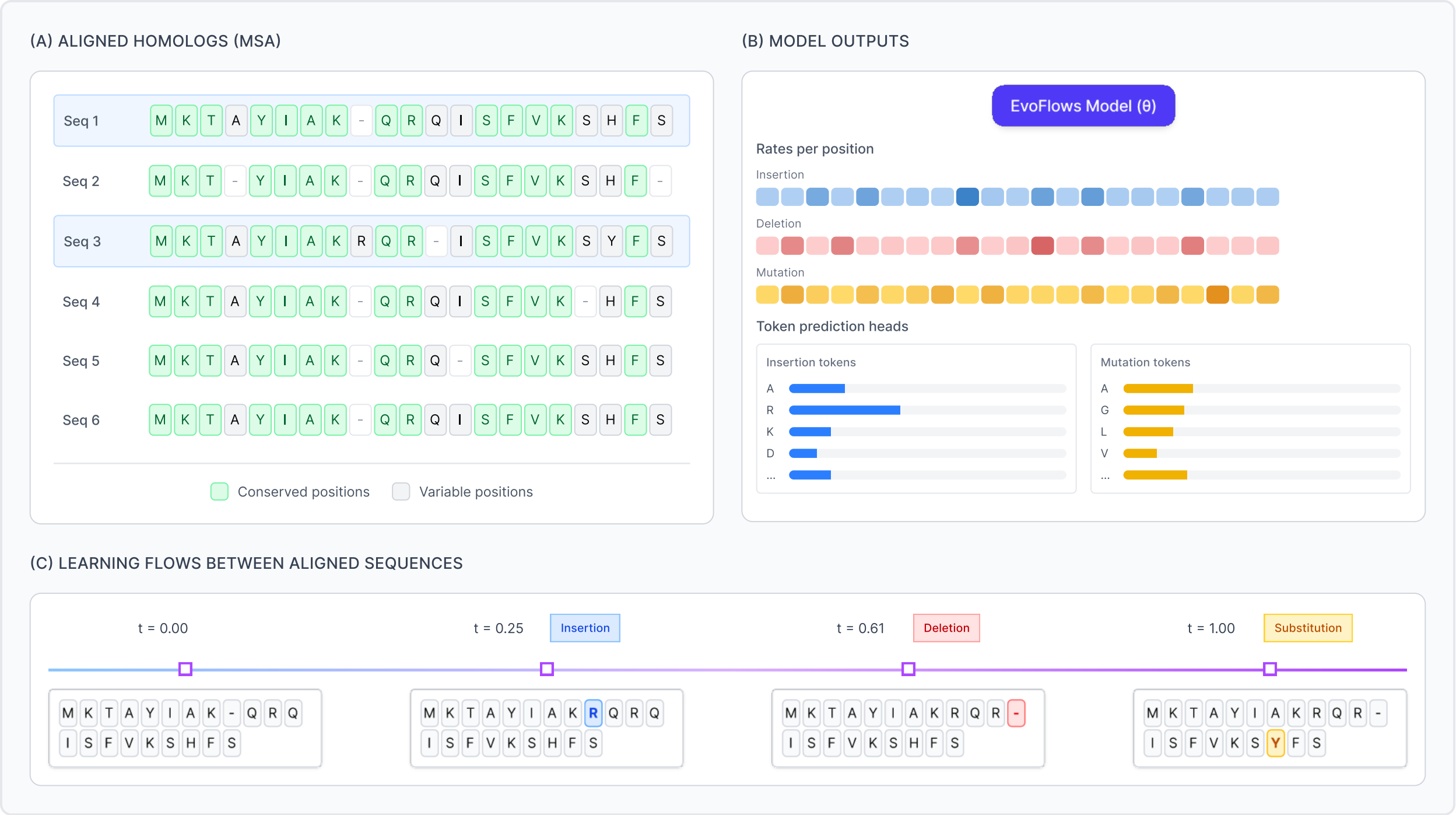

Figure 1: *Overview of EvoFlows. (A)* Homologs of a seed protein, aligned. Conserved positions are shaded green and variable positions grey; a source and target pair (highlighted) is drawn from this set to form a training pair. *(B)* The model predicts a rate for every edit type at every position, where darker means a higher predicted rate. Separate token heads predict which amino acid to use for insertions and for substitutions. *(C)* Sampling. Starting from the input sequence at $t = 0$, single edits are applied at the sampled event times until $t = 1$ returns the generated sequence.

To this end, we introduce EvoFlows, a discrete flow-matching approach (Gat et al., 2024; Havasi et al., 2025) that frames protein optimization as sequence-to-sequence editing between homologs. Rather than modeling token-level likelihoods or denoising corrupted inputs, EvoFlows generates variants by applying a controllable number of edits (substitutions, insertions, and deletions) to a template sequence (see Figure 1). We further propose a new grid-free inference procedure better suited for vectorized operations (Section 3.3). We validate this formulation on a synthetic benchmark with known ground-truth edits (Section G), and show that EvoFlows captures family-level sequence distributions competitive with state-of-the-art PLMs while generating more diverse variants without externally specified mutation sites (Section 4.2). Finally, we experimentally validate EvoFlows-generated variants of a VHH binder and an enzyme, and find that they remain functional and stable even without any computational downselection.

## 2 Preliminaries: Discrete Flow Matching and Edit Flows

A discrete flow matching model (Gat et al., 2024) learns to turn one sequence into another by making small random changes over time, with the rate of those changes trained so that the two endpoints follow a chosen joint distribution. Edit flows (Havasi et al., 2025) extend this to sequences of varying length by taking insertions and deletions, alongside substitutions, as the elementary changes.

Discrete flow matching (DFM) defines a continuous-time Markov chain (CTMC) $X_t$ over sequences of tokens in a discrete alphabet $\mathcal{A}$. DFM models approximate a joint distribution over pairs of sequences $\pi(x, x')$ by learning a local flow $u_t$ such that

$$\begin{aligned} \mathbb{P}(X_{t+h} = x_{t+h} | X_t = x_t) = \delta[x_{t+h} - x_t] + h u_t(x_{t+h}|x_t) + o(h), \\ \mathbb{P}(X_0 = x, X_1 = x') = \pi(x, x'). \end{aligned} \tag{1}$$

This permits the transport of source samples $x_0 \sim p_0(X_0)$ to target samples $x_1 \sim p_1(X_1)$ such that pairs follow the joint distribution $\pi(x_0, x_1)$.

While the original formulation of DFM was limited to fixed-length sequences, Havasi et al. (2025) introduce edit flows, a variant where transitions encoded by $u_t$ are extended to elementary edit operations (substitutions, insertions, deletions).

This target rate is generated by defining an extended space $\mathcal{Z} = \mathcal{A} \cup \{\varepsilon\}$ where $\varepsilon$ represents an empty character that enables token-wise alignment of sequence pairs $(x_0, x_1)$ by lifting them to $z_0, z_1 \in \mathcal{Z}^*$, two sequences of equal length. A conditional probability is then defined from a continuous, bijective, increasing schedule $\kappa_t : [0, 1] \to [0, 1]$

$$p_t(z|z_0, z_1) = \prod_{i \in [\![1,\ldots,|z|]\!]} \Big( (1 - \kappa_t)\, \delta_{z_0^{(i)}}\big(z^{(i)}\big) + \kappa_t\, \delta_{z_1^{(i)}}\big(z^{(i)}\big) \Big). \tag{2}$$

This leads to a conditional rate through differentiation and can be marginalized to obtain $u_t$.

Importantly, because edits are always elementary, $u_t$ can be expressed in terms of sequences in $\mathcal{A}^*$ exclusively. Given a distribution $\pi(z_0, z_1)$ that generates aligned pairs, the model $u_t^\theta$ is trained by optimizing a Bregman divergence for all times $t$:

$$\mathcal{L} = \mathop{\mathbb{E}}_{\substack{\pi(z_0,z_1) \\ t\sim\mathcal{U}(0,1) \\ p_t(z|z_0,z_1)}} \left( \sum_{x' \neq x} u_t^\theta(x'|x) - \frac{\dot{\kappa}_t}{1 - \kappa_t} \sum_{\substack{i \in [\![1,\ldots,|z|]\!] \\ z^{(i)} \neq z_1^{(i)}}} \log u_t^\theta\Big(x\Big(z, i, z_1^{(i)}\Big)|x(z)\Big) \right), \tag{3}$$

where $x(z)$ is the $\mathcal{A}^*$-sequence obtained from $z$ by removing $\varepsilon$ tokens and $x(z, i, c)$ is the $\mathcal{A}^*$-sequence obtained by inserting $c$ at position $i$ before removing $\varepsilon$. Note that the first sum is tractable because $u$ is non-zero only for pairs of sequences that differ by an elementary edit.

We find that edit flows offer a combination of features that existing PLMs do not cover: unlike MLMs, they support variable-length sequence generation without having to insert mask tokens at positions chosen in advance; and unlike autoregressive models, they modify sequences locally. We discuss existing PLMs in detail in Section A.

## 3 EvoFlows

**Problem setup.** Protein optimization starts from an existing functional protein $x_0$ and explores its local sequence neighborhood through substitutions, insertions, and deletions, while preserving overall structure and function. The task is therefore to model a conditional distribution $p(x_1 \mid x_0)$ over variants $x_1$ of the template $x_0$, rather than an unconditional distribution over sequences. This is what makes edit flows a natural fit: the flow transports $x_0$ to $x_1$ through these same elementary operations. These are also the operations by which sequences diverge in evolution, though the model learns transformations between related sequences rather than reconstructing an exact evolutionary history.

EvoFlows applies the edit-flow framework of Section 2 to protein sequences, connecting homologous proteins through a series of elementary edits: substitutions, insertions, and deletions. Unlike standard flow-matching and diffusion models, which transport a noise distribution to a data distribution, both marginals of our flow are the same distribution of natural protein homologs: EvoFlows learns to flow from one homolog to another rather than from noise to a protein.

### 3.1 Homologous protein pair distributions

An EvoFlows model should enable sampling variants of a template protein sequence $x$ that are likely to preserve its function. In the language of Section 2, this means that we should define $\pi(x, x')$ to capture some notion of $\mathbb{P}(x \text{ has a similar function to } x')$. Protein homology (meaning that two or more proteins share a common ancestor through evolution) provides a useful approximation for biochemical function similarity: related species produce similar proteins which perform analogous tasks in their respective organisms. These evolutionary relationships are established through patterns of similarity between protein sequences across species, where functional components of a protein tend to be more strongly conserved and mutations are biased toward chemically similar amino-acids (Henikoff and Henikoff, 1992).

We therefore propose the following procedure to generate the pairwise distribution $\pi(x, x')$ over pairs of protein sequences: given a starting protein $x^\dagger$, we use protein homolog search tools such as mmseqs2 (Steinegger and Söding, 2017) to query large protein databases such as UniRef30 (Suzek et al., 2007) and OAS (Olsen et al., 2022) for likely-related natural sequences. This cluster of similar protein sequences $R(x^\dagger)$ is taken as an empirical distribution over proteins that have an evolutionary relationship to $x^\dagger$:

$$p_{x^\dagger}(x) = \frac{1}{|R(x^\dagger)|} \sum_{x' \in R(x^\dagger)} \delta_{x'}(x). \tag{4}$$

So that our models capture sequence-to-sequence relationships characteristic of protein homology, we define the target $\pi(x, x') \propto \mathbb{1}(x \text{ homologous to } x')$, which we approximate as

$$\pi_{x^\dagger}(x, x') = p_{x^\dagger}(x) p_{x^\dagger}(x'). \tag{5}$$

This leads to EvoFlows learning to sample homologs $x'$ of a starting sequence $x$.

### 3.2 Pairwise protein alignment

The target path distribution is defined as an expectation over a distribution of aligned pairs $\pi(z_0, z_1)$, where $z_0$ and $z_1$ represent source and target sequences augmented with an empty token $\varepsilon$ to ensure that they have equal length. There is a natural notion of pairwise alignment for proteins, which is part of how sequence homology is estimated (Needleman and Wunsch, 1970; Smith and Waterman, 1981). This means that we can define sequence-to-sequence edit flow models in the protein space that rely on standard protein sequence alignment tools, where elementary edits including insertions and deletions correspond to actual biochemical events (Henikoff and Henikoff, 1992). We use the Needleman-Wunsch algorithm to compute alignments between pairs of protein sequences to produce augmented sequences in $\mathcal{Z}$, using global alignment (sequence-align v0.4.0) with match score +1, mismatch score −1, indel penalty −1.0, and no substitution matrix. Section C discusses how this choice shapes the conditional edit paths, and why the distribution targeted by the objective does not depend on it.

### 3.3 Inference sampling

At inference time, our generative model draws samples from $p_1$ by sampling a realization of the corresponding CTMC $x_t$ and returns $x_1 \sim p_1$:

- Initialize: Sample $x_0 \sim p_0$ and set $t_0 = 0$.
- Time-to-event: At time $t_n$, sample the duration $\Delta_n$ until the next event. Let $\tau$ denote the first-event time after $t_n$. The CDF for $\tau$ is

$$\mathbb{P}\left(\tau \leq T \mid x_{t_n}\right) = 1 - \exp\left(\int_{t_n}^{t_n+T} u_s\left(x_{t_n}|x_{t_n}\right) \mathrm{d}s\right). \tag{6}$$

  Intuitively, a higher total exit rate $-u_s\left(x_{t_n}|x_{t_n}\right)$ makes the next mutation happen sooner. We sample $\Delta_n$ via inverse CDF: draw $U \sim \text{Uniform}(0, 1)$ and numerically integrate until

$$\log(U) - \int_{t_n}^{t_n+\Delta_n} \left(u_s\left(x_{t_n}|x_{t_n}\right)\right) \mathrm{d}s = 0. \tag{7}$$

  This lets us sample the next event time directly from the total rate predicted by the model.
- Denote $t_{n+1} = t_n + \Delta_n$. If $t_{n+1} < 1$, sample

$$x_{t_{n+1}} \sim -\frac{u_{t_{n+1}}\left(\cdot\,|x_{t_n}\right)}{u_{t_{n+1}}\left(x_{t_n}|x_{t_n}\right)}, \tag{8}$$

  and continue. This step selects which specific mutation is applied next. Otherwise, return $x_{t_n}$ as $X_1$.

Note that this method is formally very close to the Euler integration proposed by Havasi et al. (2025) but replaces repeated Bernoulli variable sampling with a single uniform sampling. This approach has the practical benefit that it does not require adding potentially large sub-leading terms to (1), otherwise needed to avoid pathological distributions.

We additionally introduce a clock normalization hyperparameter that applies a length-normalized rate scaling, allowing control over the number of mutations the model performs independently of sequence length. Algorithm 1 and Algorithm 2 summarize training and generation.

### 3.4 Architecture

EvoFlows parameterizes the rate function $u_t^\theta(x' \mid x)$ using an architecture that operates directly on sequences $x \in \mathcal{A}^*$ and continuous time $t \in [0, 1]$. We use the encoder trunk of a pre-trained ESM-2 model (Lin et al., 2023) to compute a time-independent sequence embedding which is then made time-dependent using Feature-wise Linear Modulation (FiLM) (Perez et al., 2018). We finally use shallow MLPs as prediction heads to map token embeddings to rates for individual mutations. More details on the parameterization can be found in Section B. Importantly for practical applications, the late application of the FiLM time embedding means that the cost of updating the time $t$ for a fixed input is negligible. This makes inference as described in Section 3.3 efficient despite the numerical integral potentially requiring many forward passes.

# 4 Results

We evaluate EvoFlows in three stages: we first confirm on a synthetic task that the model recovers known ground-truth sequence-to-sequence edits (Section G), then benchmark it against baselines *in silico* on six natural protein homolog datasets (Section 4.2), and finally experimentally validate generated variants *in vitro* (Section 4.3).

## 4.1 Generating variants with EvoFlows

We sample $x_1 \sim p_\theta(\cdot \mid x_0)$ from a trained EvoFlows model, using the same model and inference procedure for both the *in silico* benchmark and the *in vitro* validation.

**Data and pair construction.** For the non-antibody targets, homologs are obtained via iterative profile search against UniRef30 with profile expansion and realignment, filtering for query coverage $\geq 0.8$ and E-value $\leq 10^{-1}$, and we additionally search ColabFold's environmental database to increase evolutionary depth. For the antibody fragments, we instead search the Observed Antibody Space (Olsen et al., 2022) with KA-Search (Olsen et al., 2023), which compares sequences on a fixed IMGT alignment; hits are filtered by sequence identity and coverage, then clustered by sequence identity, keeping one representative per cluster. Training pairs are formed by enumerating all unordered sequence pairs; each pair $(x_0, x_1)$ is globally aligned via Needleman-Wunsch to yield $(z_0, z_1) \in \mathcal{A}_\varepsilon^L \times \mathcal{A}_\varepsilon^L$. Dataset details are provided in Table 1.

**Inference and runtime.** We sample edits from the trained model following the procedure in Section 3.3. EvoFlows requires a number of forward passes proportional to the expected number of edits, whereas autoregressive models require one pass per token in the full sequence, and EvoDiff-MSA must repeatedly mask and regenerate all candidate edit positions.

## 4.2 *In silico* benchmark

We evaluate EvoFlows on six seed proteins: enzymes, growth factors, and antibody fragments (VHH and scFv). Each homolog set is split into train, inference, and holdout. From each inference sequence $x_0$, we generate $x_1$ with EvoFlows and five baselines, matching the expected number of mutations per sequence across methods (except random pairing). We compare the resulting $\{x_1\}$ to the holdout split as a proxy target distribution. We consider the following baselines:

**Random inference homolog pairing.** Pairs each inference sequence with a randomly selected homolog from the same protein family. Rather than generating new variants, it provides an approximate upper bound on family-level similarity, since both sequences are drawn from the natural homolog distribution.

**Evotuned PLM.** We evotune ESM-2-650M, that is, we further train it on the same homologs used for EvoFlows training (Section A). Substitution positions are sampled from a positional distribution derived from per-column entropy:

$$H(\ell) = -\sum_{a \in \mathcal{A}} p_\ell(a) \log(p_\ell(a) + \varepsilon), \tag{9}$$

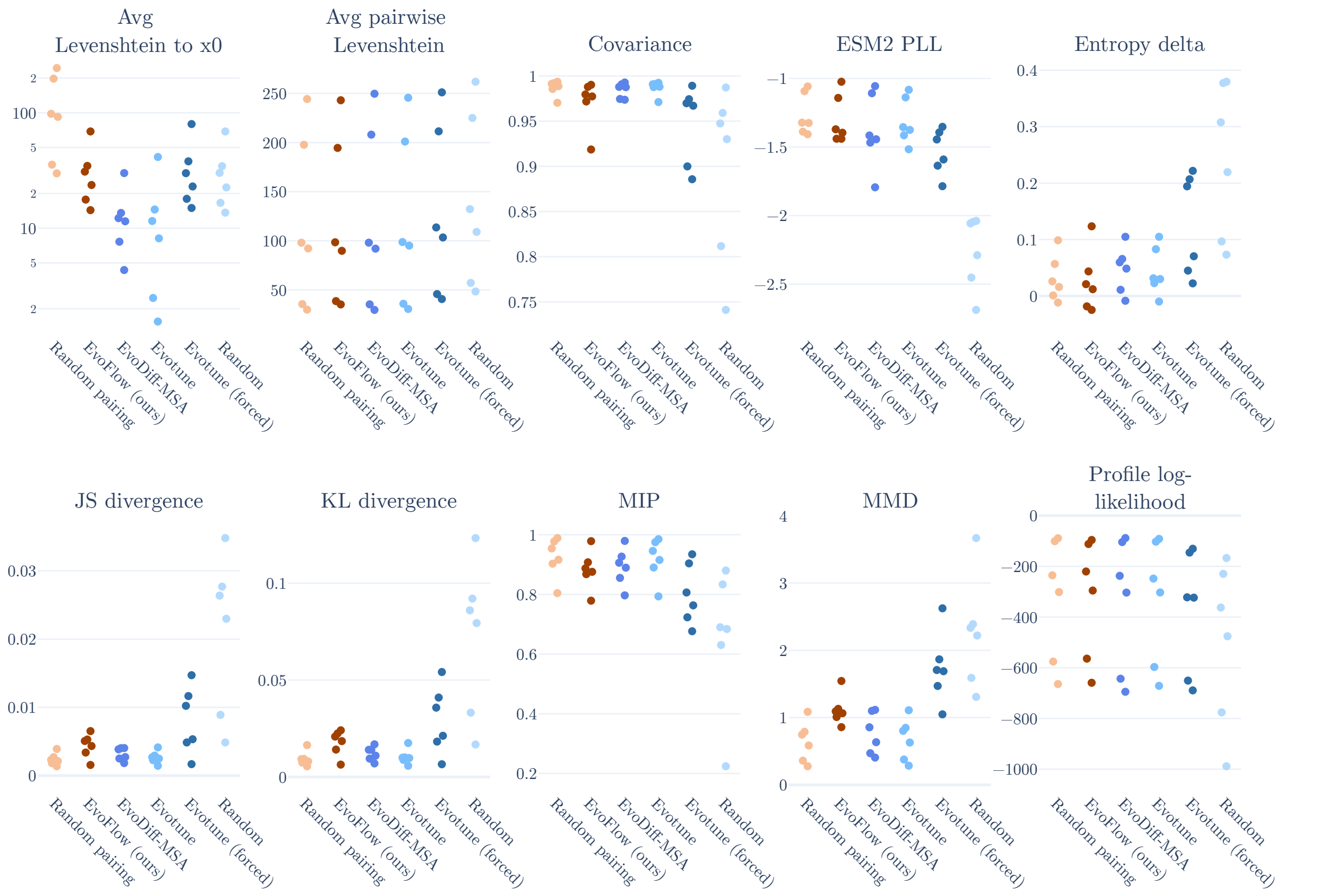

Figure 2: Comparison of EvoFlows and baseline methods across evaluation metrics, one point per homolog dataset (Section E). Higher is better for covariance, ESM2 PLL, MIP and profile log-likelihood, lower is better for entropy delta, JS and KL divergence and MMD. The two Levenshtein metrics show how far variants sit from the template and from each other, with random pairing as the natural-homolog reference. EvoFlows stays close to the holdout distribution while exploring a larger neighborhood around the template. EvoDiff-MSA is competitive on several distributional metrics but explores less, and working in aligned coordinates ties its insertions to the gap structure of the MSA: masked gap positions are often regenerated as gaps rather than amino acids, and masking whole runs of gaps makes the insertion count hard to control, whereas EvoFlows edits unaligned sequence space directly. Evotuning without forced substitutions also scores well, but only because it barely moves away from $x_0$, introducing as few as 1.5 mutations per sequence on average. Forcing substitutions raises that count, but every distributional metric gets worse. Random mutations perform worst across all metrics. Table 3 gives numerical values and compares methods at matched edit budgets, Figure 8 a per-dataset breakdown.

where $p_\ell(a)$ is the amino acid frequency in aligned column $\ell$ (excluding gaps). Entropy weights are normalized and mapped to ungapped coordinates. Masked positions are iteratively infilled using the evotuned MLM with temperature scaling.

**Evotuned PLM with forced substitutions.** Forces substitutions at masked positions by blocking infilling of the original amino acid.

**EvoDiff-MSA.** Uses the order-agnostic discrete diffusion model trained on MSA subsets (OA-DM-MAXSUB) (Alamdari et al., 2024). To generate variants, we build an MSA from the same homologs used for EvoFlows training and evotuning, then jointly align the MSA with $x_0$ using MAFFT. Mutation positions are selected using the same entropy-weighted positional distribution as the Evotuned PLM, but applied directly in aligned coordinates. That is, gapped positions can be masked, enabling insertions, and gapped tokens can be infilled, enabling deletions. Masked positions are infilled by EvoDiff-MSA conditioned on up to 64 randomly subsampled MSA sequences.

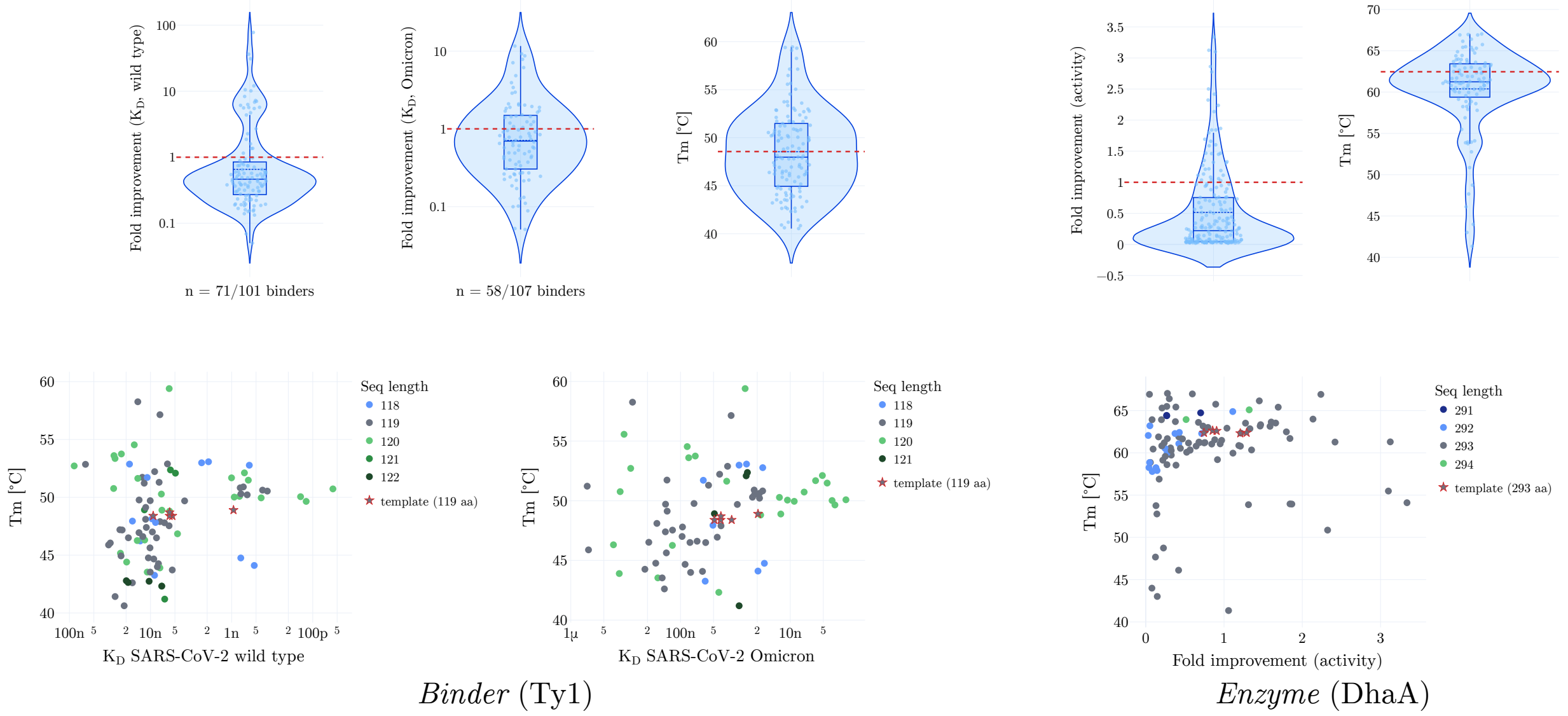

Figure 3: *In vitro validation of EvoFlows-generated variants.* Variants were expressed and assayed with no computational downselection. *(Top of each block)* One violin per assay. Binding is fold improvement over the template ($K_D$ template over $K_D$ variant, so above 1 is tighter) with non-binders excluded; the count below each $K_D$ violin is binders over sequences measured. *(Bottom of each block)* One scatter per pair of assays. Up and to the right is better in all three. Points are colored by sequence length, grey at the template length and darkening into blue for shorter variants and green for longer ones; stars mark the template. Every DhaA variant retained measurable activity with thermostability comparable to the template, and a subset improves on the template in both. Most Ty1 variants retain binding despite many edits falling in the CDR loops, and the tightest binders carry insertions and deletions.

**Random mutations.** Applies the same expected mutation count, but samples mutation positions, mutation types (substitution, insertion, deletion), and replacement amino acids uniformly, serving as a lower bound that does not preserve family-level structure.

We do not include an autoregressive baseline, since autoregressive models offer no fine-grained control over the number or location of edits, which local optimization requires. EvoFlows natively supports substitutions, insertions, and deletions, whereas the MLM baselines only support substitutions, since no widely used method exists for generating indels with MLMs.

### 4.3 *In vitro* validation

We expressed and assayed variants sampled directly from the model, with no filtering or ranking beforehand. EvoFlows never sees functional data during training, only evolutionary sequences. These experiments therefore ask whether evolutionary signal alone is enough for EvoFlows to produce functional and stable proteins. We characterized two targets (Section E): the anti-SARS-CoV-2 VHH Ty1 (binding and thermostability $T_m$) and the haloalkane dehalogenase DhaA (catalytic activity and $T_m$). Both were selected from in-house validated proteins with established assays, chosen to span two protein classes and two functional readouts. Assay protocols are given in Section F.

**Binder.** Of 177 designed Ty1 variants, carrying 1-10 edits each (Figure 6), 100 yielded a binding measurement against the SARS-CoV-2 wild type and 106 against the Omicron variant. Including the template, 71 of 101 (70%) bound the wild type and 58 of 107 (54%) Omicron (Figure 3). Many of the edits fall in the CDR loops that form the paratope (Figure 6), so with no assay data and no filtering, some loss of binding is expected. Notably, the variants that bind more tightly than the template carry insertions and deletions, reaching a part of sequence space that substitution-only methods cannot.

**Enzyme.** All 173 designed DhaA variants, carrying 1-13 edits each (Figure 7), retained measurable catalytic activity. Thermostability was measured for 85 of these and remained comparable to the template; 26 of those 85 (31%) also exceeded the catalytic activity of the template, the most active by 3.3-fold (Figure 3).

## 5 Conclusion

We introduced EvoFlows, an edit-based discrete flow matching framework that learns protein transition rules through substitutions, insertions, and deletions. To our knowledge, EvoFlows is the first method to model insertions and deletions as explicit edit operations in a template-based protein optimization setting, a capability that existing masked, autoregressive, and diffusion protein language models lack.

We extend the original edit flow framework in two ways. First, protein sequences are a particularly natural fit: unlike text or code, protein evolution operates through discrete, well-defined mutations, and large datasets of aligned protein families are readily available for training. Second, we introduce a deterministic benchmark that, for the first time, validates that edit flows can reliably learn local transition distributions in a sequence-to-sequence setting. Across six diverse protein families spanning enzymes, growth factors, and antibody fragments, EvoFlows generates non-trivial, natural-like variants that explore substantially larger mutational neighborhoods than state-of-the-art baselines while remaining close to the holdout distribution.

Beyond these *in silico* results, we experimentally characterized EvoFlows-generated variants of the Ty1 VHH and the DhaA enzyme. Sampling directly from the model without computational downselection, all 173 assayed DhaA variants retained activity and 70% of Ty1 variants retained binding, including variants carrying insertions and deletions, despite the model never having seen a functional measurement.

**Limitations.** EvoFlows is family specific by design: a model is trained on homologs of the protein being optimized, which fits lead optimization but means a new family requires a new model. Homolog pairs are sampled uniformly, so densely represented subfamilies contribute more training pairs and the distribution of evolutionary distances is not explicitly controlled. It learns a prior over plausible local sequence changes rather than directly optimizing for particular engineering objectives. Our benchmark also covers only six protein families.

**Outlook.** We see train-time and inference-time conditioning as promising directions for practical protein engineering: property-based guidance, sequence constraints (motifs, mutation budgets), and metadata conditioning (species, gene ontology). Such mechanisms would enable EvoFlows to serve as a foundation for guided protein optimization.

## Acknowledgments

We thank the Cradle wet lab team for experimental support.

## References

Alamdari, S., Thakkar, N., Berg, R. van den, Tenenholtz, N., Strome, R., Moses, A. M., Lu, A. X., Fusi, N., Amini, A. P., and Yang, K. K. Protein generation with evolutionary diffusion: sequence is all you need. *Biorxiv*, 2024. https://doi.org/10.1101/2023.09.11.556673

Alley, E. C., Khimulya, G., Biswas, S., AlQuraishi, M., and Church, G. M. Unified rational protein engineering with sequence-based deep representation learning. *Nature Methods*, *16*(12), 1315–1322, 2019.

Altschul, S. F., Gertz, E. M., Agarwala, R., Schäffer, A. A., and Yu, Y.-K. PSI-BLAST pseudocounts and the minimum description length principle. *Nucleic Acids Research*, *37*(3), 815–824, 2009. https://doi.org/10.1093/nar/gkn981

Austin, J., Johnson, D. D., Ho, J., Tarlow, D., and Berg, R. van den. Structured Denoising Diffusion Models in Discrete State-Spaces. *Advances in Neural Information Processing Systems*, *34*, 17981–17993, 2021. https://proceedings.neurips.cc/paper_files/paper/2021/file/958c530554f78bcd8e97125b70e6973d-Paper.pdf

Brandes, N., Ofer, D., Peleg, Y., Rappoport, N., and Linial, M. ProteinBERT: a universal deep-learning model of protein sequence and function. *Bioinformatics*, *38*(8), 2102–2110, 2022. https://doi.org/10.1093/bioinformatics/btac020

BRENDA Enzyme Database. *Information on EC 2.6.1.51 - serine-pyruvate transaminase*, 2025.

Chen, A., Stanton, S. D., Alberstein, R. G., Watkins, A. M., Bonneau, R., Gligorijevic, V., Cho, K., and Frey, N. C. LLMs Are Highly-Constrained Biophysical Sequence Optimizers. *NeurIPS 2024 Workshop on AI for New Drug Modalities*, 2024, December.

Dockalova, V., Sanchez-Carnerero, E. M., Dunajova, Z., Palao, E., Slanska, M., Buryska, T., Damborsky, J., Klán, P., and Prokop, Z. Fluorescent substrates for haloalkane dehalogenases: Novel probes for mechanistic studies and protein labeling. *Computational and Structural Biotechnology Journal*, *18*, 922–932, 2020. https://doi.org/10.1016/j.csbj.2020.03.029

Dunn, S., Wahl, L., and Gloor, G. Mutual information without the influence of phylogeny or entropy dramatically improves residue contact prediction. *Bioinformatics*, *24*(3), 333–340, 2007. https://doi.org/10.1093/bioinformatics/btm604

Elnaggar, A., Heinzinger, M., Dallago, C., Rehawi, G., Wang, Y., Jones, L., Gibbs, T., Feher, T., Angerer, C., Steinegger, M., Bhowmik, D., and Rost, B. ProtTrans: Toward Understanding the Language of Life Through Self-Supervised Learning. *IEEE Transactions on Pattern Analysis and Machine Intelligence*, *44*(10), 7112–7127, 2022. https://doi.org/10.1109/tpami.2021.3095381

Ferruz, N., Schmidt, S., and Höcker, B. ProtGPT2 is a deep unsupervised language model for protein design. *Nature Communications*, *13*(1), 4348, 2022. https://doi.org/10.1038/s41467-022-32007-7

Gat, I., Remez, T., Shaul, N., Kreuk, F., Chen, R. T., Synnaeve, G., Adi, Y., and Lipman, Y. Discrete flow matching. *Advances in Neural Information Processing Systems*, *37*, 133345–133385, 2024.

Gretton, A., Borgwardt, K. M., Rasch, M. J., Schölkopf, B., and Smola, A. A Kernel Two-Sample Test. *Journal of Machine Learning Research*, *13*(25), 723–773, 2012. http://jmlr.org/papers/v13/gretton12a.html

Gruver, N., Stanton, S. D., Frey, N. C., Rudner, T. G. J., Hotzel, I., Lafrance-Vanasse, J., Rajpal, A., Cho, K., and Wilson, A. G. Protein Design with Guided Discrete Diffusion. *Thirty-Seventh Conference on Neural Information Processing Systems*, 2023. https://openreview.net/forum?id=MfiK69Ga6p

Hallee, L., Rafailidis, N., Bichara, D. B., and Gleghorn, J. P. *Diffusion Sequence Models for Enhanced Protein Representation and Generation*, 2025. https://arxiv.org/abs/2506.08293

Hanke, L., Vidakovics Perez, L., Sheward, D. J., Das, H., Schulte, T., Moliner-Morro, A., Cber, M., Vlasova, N., Schul, S., and others. An alpaca nanobody neutralizes SARS-CoV-2 by blocking receptor interaction. *Nature Communications*, *11*(1), 4420, 2020. https://doi.org/10.1038/s41467-020-18174-5

Havasi, M., Karrer, B., Gat, I., and Chen, R. T. Edit Flows: Variable Length Discrete Flow Matching with Sequence-Level Edit Operations. *The Thirty-Ninth Annual Conference on Neural Information Processing Systems*, 2025.

Henikoff, S., and Henikoff, J. G. Amino acid substitution matrices from protein blocks. *Proceedings of the National Academy of Sciences*, *89*(22), 10915–10919, 1992. https://doi.org/10.1073/pnas.89.22.10915

Hie, B. L., Shanker, V. R., Xu, D., Bruun, T. U. J., Weidenbacher, P. A., Tang, S., Wu, W., Pak, J. E., and Kim, P. S. Efficient Evolution of Human Antibodies from General Protein Language Models. *Nature Biotechnology*, *42*(2), 275–283, 2024. https://doi.org/10.1038/s41587-023-01763-2

Hoogeboom, E., Gritsenko, A. A., Bastings, J., Poole, B., Berg, R. van den, and Salimans, T. Autoregressive Diffusion Models. *International Conference on Learning Representations*, 2022. https://openreview.net/forum?id=Lm8T39vLDTE

Jr, T. F. T., and Bepler, T. Understanding protein function with a multimodal retrieval-augmented foundation model. *The Thirty-Ninth Annual Conference on Neural Information Processing Systems*, 2025. https://openreview.net/forum?id=fKerD2AQai

Koudelakova, T., Bidmanova, S., Dvorak, P., Pavelka, A., Chaloupkova, R., Prokop, Z., and Damborsky, J. Haloalkane dehalogenases: biotechnological applications. *Biotechnology Journal*, *8*(1), 32–45, 2013. https://doi.org/10.1002/biot.201100486

Koçer, İ., Cox, E. C., DeLisa, M. P., and Bhattacharya, R. Effects of variable domain orientation on anti-HER2 single-chain variable fragment antibody expressed in the Escherichia coli cytoplasm. *Biotechnology Progress*, *37*(2), e3102, 2021. https://doi.org/10.1002/btpr.3102

Kucera, T., Togninalli, M., and Meng-Papaxanthos, L. Conditional generative modeling for de novo protein design with hierarchical functions. *Bioinformatics*, *38*(13), 3454–3461, 2022. https://doi.org/10.1093/bioinformatics/btac353

Leslie, C., Eskin, E., and Noble, W. S. The spectrum kernel: a string kernel for SVM protein classification. *Pac Symp Biocomput*, 564–575, 2002.

Lin, Z., Akin, H., Rao, R., Hie, B., Zhu, Z., Lu, W., Smetanin, N., Verkuil, R., Kabeli, O., Shmueli, Y., and others. Evolutionary-scale prediction of atomic-level protein structure with a language model. *Science*, *379*(6637), 1123–1130, 2023.

Listov, D., Goverde, C. A., Correia, B. E., and Fleishman, S. J. Opportunities and challenges in design and optimization of protein function. *Nature Reviews Molecular Cell Biology*, *25*(8), 639–653, 2024.

Madani, A., Krause, B., Greene, E. R., Subramanian, S., Mohr, B. P., Holton, J. M., Olmos, J. L., Xiong, C., Sun, Z. Z., Socher, R., Fraser, J. S., and Naik, N. Large language models generate functional protein sequences across diverse families. *Nature Biotechnology*, *41*(8), 1099–1106, 2023. https://doi.org/10.1038/s41587-022-01618-2

Munsamy, G., Illanes-Vicioso, R., Funcillo, S., Nakou, I. T., Lindner, S., Ayres, G., Sheehan, L. S., Moss, S., Eckhard, U., Lorenz, P., and Ferruz, N. *Conditional Language Models Enable the Efficient Design of Proficient Enzymes* (p. 2024.05.03.592223). bioRxiv, 2024. https://doi.org/10.1101/2024.05.03.592223

Needleman, S. B., and Wunsch, C. D. A general method applicable to the search for similarities in the amino acid sequence of two proteins. *Journal of Molecular Biology*, *48*(3), 443–453, 1970. https://doi.org/10.1016/0022-2836(70)90057-4

Olsen, T. H., Abanades, B., Moal, I. H., and Deane, C. M. KA-Search, a method for rapid and exhaustive sequence identity search of known antibodies. *Scientific Reports*, *13*(1), 11612, 2023. https://doi.org/10.1038/s41598-023-38108-7

Olsen, T. H., Boyles, F., and Deane, C. M. Observed Antibody Space: A diverse database of cleaned, annotated, and translated unpaired and paired antibody sequences. *Protein Science*, *31*(1), 141–146, 2022. https://doi.org/10.1002/pro.4205

Paul, S. M., Mytelka, D. S., Dunwiddie, C. T., Persinger, C. C., Munos, B. H., Lindborg, S. R., and Schacht, A. L. How to Improve R&D Productivity: The Pharmaceutical Industry's Grand Challenge. *Nature Reviews Drug Discovery*, *9*(3), 203–214, 2010. https://doi.org/10.1038/nrd3078

Perez, E., Strub, F., De Vries, H., Dumoulin, V., and Courville, A. Film: Visual reasoning with a general conditioning layer. *Proceedings of the AAAI Conference on Artificial Intelligence*, *32*(1), 2018.

Rock, E. P., Sibbald, P. R., Davis, M. M., and Chien, Y. H. CDR3 length in antigen-specific immune receptors. *Journal of Experimental Medicine*, *179*(1), 323–328, 1994. https://doi.org/10.1084/jem.179.1.323

Roovers, R. C., Vosjan, M. J., Laeremans, T., Khoulati, R. el, Bruin, R. C. de, Ferguson, K. M., Lemmon, M. A., Dongen, G. A. van, and Henegouwen, P. M. van Bergen en. Efficient inhibition of EGFR signaling and of tumour growth by antagonistic anti-EGFR nanobodies. *Cancer Immunology, Immunotherapy*, *60*(8), 1047–1058, 2011. https://doi.org/10.1007/s00262-011-1027-2

Smith, T. F., and Waterman, M. S. Identification of common molecular subsequences. *Journal of Molecular Biology*, *147*(1), 195–197, 1981. https://doi.org/10.1016/0022-2836(81)90087-5

Steinegger, M., and Söding, J. MMseqs2 enables sensitive protein sequence searching for the analysis of massive data sets. *Nature Biotechnology*, *35*(11), 1026–1028, 2017. https://doi.org/10.1038/nbt.3988

Suzek, B. E., Huang, H., McGarvey, P., Mazumder, R., and Wu, C. H. UniRef: comprehensive and non-redundant UniProt reference clusters. *Bioinformatics*, *23*(10), 1282–1288, 2007. https://doi.org/10.1093/bioinformatics/btm098

UniProt Consortium. *FGF2 - Fibroblast growth factor 2 - Gallus gallus (Chicken)*, 2025.

Verkuil, R., Kabeli, O., Du, Y., Wicky, B. I. M., Milles, L. F., Dauparas, J., Baker, D., Ovchinnikov, S., Sercu, T., and Rives, A. Language models generalize beyond natural proteins. *Biorxiv*, 2022. https://doi.org/10.1101/2022.12.21.521521

Widatalla, T., Rafailov, R., and Hie, B. *Aligning Protein Generative Models with Experimental Fitness via Direct Preference Optimization*, 2024. https://doi.org/10.1101/2024.05.20.595026

# A Extended Related Work

## A.1 Protein language models (PLMs)

Protein language models (PLMs) learn representations of protein sequences by pre-training on large corpora of unlabeled sequences. Common pre-training methods include masked language modeling (MLM) (Brandes et al., 2022; Elnaggar et al., 2022; Verkuil et al., 2022), autoregressive PLMs (Chen et al., 2024; Ferruz et al., 2022; Jr and Bepler, 2025; Madani et al., 2023), and discrete diffusion-based sequence models (Alamdari et al., 2024; Hallee et al., 2025). These models have been shown to capture structural, functional, and evolutionary information, and serve as foundations for downstream prediction and generation tasks.

**Autoregressive PLMs** parameterize a distribution over protein sequences as

$$p_\theta(x) = \prod_{i=1}^{L} p_\theta(x_i \mid x_{<i}), \tag{10}$$

where $x = (x_1, ..., x_L) \in \mathcal{A}^L$ is a protein sequence over the amino acid alphabet $\mathcal{A}$, and each conditional probability $p_\theta(x_i \mid x_{<i})$ is a categorical distribution over amino acids. We denote $x_{<i}$ as a sequence where positions $\geq i$ are masked.

Autoregressive models have been successful for *de novo*-like tasks such as enzyme generation with weak conditioning (Madani et al., 2023; Munsamy et al., 2024). However, they generate entire sequences, making it difficult to control where mutations occur and how many mutations are introduced, making them less suitable for protein optimization tasks that only require a few targeted mutations. Additionally, methods generating local variants rely on conditioning on one (Chen et al., 2024) or multiple context sequences (Jr and Bepler, 2025) but require the model to generate the entire sequence to obtain a few mutations, which is inefficient.

**Masked PLMs** are often perceived as a more parsimonious approach for protein variant generation because they directly edit a subset of positions while preserving the rest of the sequence. MLMs instead learn the conditional distributions of the form

$$p_\theta\left(x_i | x_{\setminus i}\right), \quad i = 1, ..., L, \tag{11}$$

where $x_{\setminus i}$ denotes the sequence with position $i$ masked. As in the autoregressive case, the model outputs a categorical distribution over amino acids at each position.

Masked PLMs define residue-level conditional distributions rather than a distribution over protein sequences. While they can be used for sequence generation through iterative decoding, this requires specifying mask locations and assumes a fixed sequence length.

**Discrete diffusion models** generalize masked and autoregressive language modeling by introducing an explicit generation path through a sequence of intermediate states. They define a stochastic forward corruption process over discrete sequences and learn to invert this process through denoising (Austin et al., 2021; Hoogeboom et al., 2022).

Formally, a diffusion model defines a forward Markov process that progressively corrupts a clean sequence $x^{(0)}$ and learns the reverse process

$$p_\theta\left(x^{(t-1)} | x^{(t)}\right), \tag{12}$$

which iteratively denoises toward the data distribution.

In protein diffusion models, the corruption is often defined by iteratively introducing "absorbing" states ([MASK] tokens)(Alamdari et al., 2024) to which valid tokens are converted. This framework is a generalization that encompasses both autoregressive (left-to-right denoising paths) and MLM (single step in order-agnostic paths).

This generalization also appears in potential applications of diffusion models in protein optimization: conditioning on a starting sequence requires applying some level of corruption, which raises the same issue for picking *where* that corruption should be applied like for masked PLMs. On the other hand, starting from a fully corrupted state requires the same level of unnecessary computation and conditioning as autoregressive models.

### A.2 Evotuning

Evotuning (Alley et al., 2019) adapts pre-trained PLMs to a specific protein family by further training on sequences drawn from a homologous sequence set, such as those retrieved from a protein database by sequence-similarity search (Steinegger and Söding, 2017), using the same self-supervised learning objective. Evotuned models have been shown to better capture family-specific constraints and improve performance on downstream prediction and design tasks (Alley et al., 2019). Given the opportunity for improvement on tasks limited to a single protein family, foundational PLMs should be evotuned when compared to family-specific models like EvoFlows.

## B Architecture Details

**Sequence embedding:** Given a sequence $x_t \in \mathcal{A}^L$, we compute token embeddings using a pretrained ESM-2 model,

$$h_t = \mathrm{ESM2}(x_t) \in \mathbb{R}^{L \times D}, \tag{13}$$

where $D$ is the dimension of the embeddings. To maintain consistency with ESM-2, the embedder includes explicit start and end tokens. During training, the ESM-2 encoder weights are finetuned jointly with the EvoFlows model rather than frozen.

**Time conditioning:** We embed the time $t$ using a sinusoidal embedding followed by an MLP:

$$\tau_t = \mathrm{MLP}(\mathrm{Sinusoidal}(t)) \in \mathbb{R}^D. \tag{14}$$

This time embedding is used to condition the token representations via Feature-wise Linear Modulation (FiLM) (Perez et al., 2018), which learns a feature-wise affine transformation that maps $\tau_t$ to scale and shift vectors $(\gamma_t, \beta_t) \in \mathbb{R}^D$. The token embeddings are transformed as

$$\tilde{h}_t[i] = h_t[i] \odot (1 + \gamma_t) + \beta_t, \tag{15}$$

and are shared across all heads.

Notably, the proposed time-conditioning mechanism enables efficient changing of time for fixed inputs, which is particularly useful for training with multiple time samples per sequence in a batch.

**Rate prediction:** The rate $u_t^\theta(x'|x)$ is obtained through a similar parameterization as what is adopted by Havasi et al. (2025):

$$u_t^\theta(x'|x) = \begin{cases} \lambda_t^{\text{sub}}\big(\tilde{h}_t[i]\big)Q_t^{\text{sub}}\big(\tilde{h}_t[i], i, a\big) \text{ if } x' = \text{sub}(x, i, a) \\ \lambda_t^{\text{ins}}\big(\tilde{h}_t[i]\big)Q_t^{\text{ins}}\big(\tilde{h}_t[i], i, a\big) \text{ if } x' = \text{ins}(x, i, a) \\ \lambda_t^{\text{del}}\big(\tilde{h}_t[i]\big) \text{ if } x' = \text{del}(x, i) \\ 0 \text{ otherwise} \end{cases}, \tag{16}$$

where $\text{sub}(x, i, a)$ denotes $x$ transformed by a substitution by the token $a$ at position $i$, $\text{ins}(x, i, a)$ denotes $x$ transformed by an insertion of the token $a$ after position $i$, and $\text{del}(x, i)$ denotes $x$ transformed by a deletion at position $i$.

Both $\lambda_t^\bullet$ and $Q_t^\bullet$ are defined as shallow MLPs and denote respectively positional rates and distributions over tokens. Deletions carry no token distribution, so their rate is $\lambda_t^{\text{del}}$ alone. The positional rates $\lambda_t^\bullet$ are normalized using either a softplus transformation or a bounded sigmoid mapping to ensure the rates are positive and numerically stable. The token distributions $Q_t^\bullet$ are parameterized as ESM2 language prediction heads.

**Insertion slots.** Substitutions and deletions act on existing residues, so their rates are predicted from token representations and one rate is emitted per position. Insertions instead act between residues, so they are predicted from gap representations: for a sequence of $L$ tokens there are $L + 1$ gaps, and the representation of gap $c$ is the mean of the modulated embeddings $\tilde{h}_t[c-1]$ and $\tilde{h}_t[c]$, falling back to the single available neighbour at the two ends. The insertion head therefore emits $L + 1$ rates, one per slot between tokens, while the deletion and substitution rates are emitted by a single token-level head with two outputs per position.

**Token heads.** The insertion and substitution distributions $Q_t^\bullet$ are initialized as copies of the pretrained ESM-2 language modelling head, and both use the backbone input embedding matrix as a tied decoder weight, so that the decoder stays consistent with the trunk while it is finetuned. Deletions carry no distribution over tokens, since the operation is fully determined by its position.

**Training and sampling procedures.** Algorithm 1 and Algorithm 2 summarize training and generation. Training samples an interpolant between the two aligned sequences, while generation walks the template forward one edit at a time until $t = 1$.

**input:** homolog set $\mathcal{H}$, model $u^\theta$
sample a pair $(x_0, x_1)$ from $\mathcal{H}$
$(z_0, z_1) \leftarrow \text{align}(x_0, x_1)$ *(equal length, gapped)*
sample $t \sim \text{Uniform}(0, 1)$
**for** each aligned position $j$ **do**
  $z_t[j] \leftarrow z_1[j]$ with probability $\kappa(t) = t$, else $z_0[j]$
**end**
$x_t \leftarrow$ remove gaps from $z_t$
$(\lambda, Q) \leftarrow u^\theta(x_t, t)$
$\mathcal{L} \leftarrow \sum_e \lambda_e - \frac{\dot{\kappa}(t)}{1-\kappa(t)} \sum_{\text{edits remaining in } z_1} \log(\lambda_e Q_e)$
$\theta \leftarrow \theta - \eta \nabla_\theta \mathcal{L}$

Algorithm 1: EvoFlows training step.

Each position switches from $z_0$ to $z_1$ independently, so no ordering of the edits is privileged, as discussed in Section C.

**input:** template $x_0$, model $u^\theta$, step size $dt$
$x \leftarrow x_0$, $t \leftarrow 0$
**while** $t < 1$ **do**
  $h \leftarrow \mathrm{PLM}(x)$ *(independent of $t$, recomputed only after an edit)*
  sample $u \sim \mathrm{Uniform}(0, 1)$, set $s \leftarrow \log(1 - u)$
  **while** $s < 0$ **and** $t < 1$ **do** *(inverse CDF by accumulation)*
    $s \leftarrow s + R(h, t) \cdot dt$, where $R$ is the total rate over all edits
    $t \leftarrow t + dt$
  **end**
  **if** $t < 1$ **then**
    $(\lambda, Q) \leftarrow \mathrm{head}(h, t)$
    sample an edit type and position $(e, i)$ with probability proportional to $\lambda$
    **if** $e$ is a substitution or an insertion **then** sample $a \sim Q_e(\cdot, i)$ **end**
    apply the edit to $x$
  **end**
**end**
**return** $x$

Algorithm 2: Sampling a variant $x_1$ from a template $x_0$.

## C Alignment Choice and Edit Paths

**Dependence on the alignment.** The method depends on the chosen alignment only through the coupling between $x_0$ and $x_1$. The alignment determines which residues correspond to one another, and therefore whether a given difference is represented as a substitution or as a deletion together with an insertion. Because Needleman-Wunsch optimizes a scoring function, and several alignments may be optimal under it, different valid alignments can define different conditional edit paths for the same pair of sequences. However, every such coupling has the same source and target marginals, so changing the alignment does not change the distribution that the flow-matching objective targets. It changes the paths used to learn that distribution, not the distribution itself. This invariance holds at the optimum, so a trained model may still retain some sensitivity to the alignment used.

**Equivalent orderings of edits.** For a given alignment, all orderings of the resulting edits are treated equivalently. The alignment represents the pair as two gapped sequences of equal length, in which each differing column corresponds to one edit. Our path assigns each of these edits an independent switch time, so training averages over orderings rather than committing to a particular sequence of operations. The state $x_t$ records only which edits have occurred by time $t$, not the order in which they were applied. A different alignment can instead define a different set of edits, which is the case discussed above.

## D Evaluation Metrics

In this section, we describe the selected metrics for sequence and dataset distribution evaluation and comparison in detail. Generative model evaluation has been notoriously difficult, and we aim to choose metrics and distances suitable for the proposed task of protein sequence generation.

### D.1 Dataset Heuristics

We hypothesize that a well performing, unconditioned generation should be distributionally indistinguishable from a subsampled holdout set of the multiple sequence alignment used for training. The generative process should thus recreate this set of defined heuristics. At a sequence level, we consider the distribution of sequence lengths, the overall distribution of amino acid residues, and the per-position distribution of amino acids and gaps with respect to a fixed sequence alignment.

### D.2 Model-Based Pseudo Log Likelihood

It is possible to quantify the likelihood of a sequence under a probabilistic model. The most commonly known method related to LLMs is perplexity (PPL, applicable to autoregressive models), or pseudo log likelihood (PLL, applicable to masked language models). For both models the cumulative likelihood of a generative path is calculated. While AR models can be traversed in a single, well-defined path, this does not hold true for MLM. For the latter, we choose a single random order over individually masked positions to calculate the pseudo log likelihood.

### D.3 Covariance and Mutual Information

A strong indicator for models correctly capturing signals connected to coevolution is the proper covariance of individual positions in an alignment. Covariance can be an indicator of sequences that co-evolved with epistatic changes, for instance to maintain structural contacts.

In natural protein evolution, amino acid substitutions do not occur independently. When a mutation changes one position in a protein structure, compensatory mutations at other positions often follow to maintain protein stability and function. These co-evolving positions exhibit statistical dependencies that can be detected as covariance in multiple sequence alignments. A generative model that accurately captures these evolutionary constraints should reproduce the covariance patterns observed in natural protein families.

**4D Covariance.** Given a set of one-hot-encoded sequences $\mathcal{X} = \{x_1, ..., x_N\} \subset \{0,1\}^{\mathtt{L}\times 21}$ from a joint alignment, we define the positional amino acid frequency $f_i(a)$, the joint frequency $f_{ij}(a,b)$, and the covariance $C_{ij}(a,b)$ as:

$$\begin{aligned} f_i(a) &= \frac{1}{N}\sum_n^N x_{nia} \in \mathbb{R}^{\mathtt{L}\times 21}, \\ f_{ij}(a,b) &= \frac{1}{N}\sum_n^N x_{nia}x_{njb} \in \mathbb{R}^{\mathtt{L}\times\mathtt{L}\times 21\times 21}, \\ C_{ij}(a,b) &= f_{ij}(a,b) - f_i(a)f_j(b) \in \mathbb{R}^{\mathtt{L}\times\mathtt{L}\times 21\times 21}. \end{aligned} \tag{17}$$

**Positional Interaction Strength.** The 4-dimensional tensor $C_{ij}(a,b)$ captures the per-position, per-amino-acid covariance over a dataset. We can contract this information by using the Frobenius norm over the last two dimensions, which can be interpreted as contracting the per-residue coupling into an overall interaction strength at a given pair of positions $i$ and $j$:

$$\|C_{ij}\|_F = \sqrt{\sum_{a,b} C_{ij}(a,b)^2}. \tag{18}$$

**Mutual Information with Average Product Correction.** An interpretable 2D representation of coevolution can also be constructed from information theory, as described by (Dunn et al., 2007). Using the definitions from equation (17), we can describe the mutual information $I_{ij}$ as:

$$I_{ij} = \sum_{a,b} f_{ij}(a,b) \log \frac{f_{ij}(a,b)}{f_i(a) f_j(b)}. \tag{19}$$

We denote column, row, and global means as

$$I_i = \frac{1}{L} \sum_{k=1}^{L} I_{ik}, \quad I_j = \frac{1}{L} \sum_{k=1}^{L} I_{kj}, \quad I = \frac{1}{L^2} \sum_{i,j} I_{ij}, \tag{20}$$

and define the Average Product Correction $\mathrm{APC}_{ij}$ as

$$\mathrm{APC}_{ij} = \frac{I_i I_j}{I}. \tag{21}$$

This lets us define MIp, the mutual information with average product correction, as

$$\mathrm{MIp}_{ij} = I_{ij} - \mathrm{APC}_{ij}. \tag{22}$$

### D.4 BLOSUM-Corrected KL Divergence

While the metrics described above characterize properties of individual datasets, this section focuses on distance measures that quantify the similarity between two datasets. These measures are essential for comparing synthetic sequences generated by our models against held-out natural sequences from the training MSA. The Kullback-Leibler (KL) divergence provides a principled information-theoretic approach to measure how one probability distribution differs from another.

Given two discrete probability distributions $p$ and $q$ over amino acid frequencies, the KL divergence is defined as:

$$D_{\mathtt{KL}}(p \parallel q) = \sum_{a} p(a) \log \frac{p(a)}{q(a)}, \tag{23}$$

where the sum is taken over all amino acid types. The KL divergence is always non-negative and equals zero if and only if $p$ and $q$ are identical distributions. However, $D_{\mathtt{KL}}$ is asymmetric: $D_{\mathtt{KL}}(p \parallel q) \neq D_{\mathtt{KL}}(q \parallel p)$ in general.

**The Zero-Frequency Problem and Biological Priors.** A critical challenge in computing KL divergence for protein sequences arises from the zero-frequency problem. In small datasets or at highly conserved positions, certain amino acids may not appear at all, leading to zero probabilities. When $q(a) = 0$ and $p(a) > 0$, the KL divergence becomes infinite, making comparisons impossible.

Additive smoothing (also known as Lidstone smoothing) addresses this by adding pseudo-counts to observed frequencies. Rather than using uninformative uniform pseudo-counts, such as `1/vocabulary_size`, we employ `BLOSUM62` background frequencies (Altschul et al., 2009) as biologically informed priors. These frequencies reflect the natural abundance of amino acids in protein databases and have been empirically validated across diverse protein families. The smoothed probability for amino acid $i$ is calculated as:

$$p_{a,\alpha} = \frac{x_a + \alpha\mu_a}{N + \alpha \cdot d}. \tag{24}$$

where $x_a$ is the observed count of amino acid $a$, $N$ is the total number of observations, $d = 21$ is the number of possible amino acids (20 standard amino acids plus gap), $\mu_a$ is the BLOSUM62 prior frequency for amino acid $a$, and $\alpha$ is the pseudo-count weight parameter. We calibrated the gap frequency $\mu_{\texttt{GAP}}$ empirically across all MSAs used in our evaluation, finding a value of 0.7% to match most common cases using a coverage-based proxy for pairwise alignment sparsity.

**Biological Justification.** The use of BLOSUM62 frequencies as priors is biologically motivated. These frequencies were derived from the BLOCKS database of conserved protein regions and reflect amino acid propensities averaged over many protein families. By incorporating this knowledge, our smoothing method:

1. Ensures biologically plausible probability estimates even with limited data
2. Weights pseudo-counts according to amino acid abundance in natural proteins rather than treating all amino acids equally
3. Provides a Bayesian interpretation with Dirichlet priors informed by empirical protein evolution
4. Reduces the variance of frequency estimates while introducing minimal bias, as the smoothed probabilities converge to maximum likelihood estimates as dataset size increases

This approach is particularly valuable when comparing generated sequences to natural sequences, as it ensures that distance measurements remain finite and meaningful even when datasets differ in size or sampling depth, while respecting the underlying biochemical constraints of protein composition.

### D.5 Spectrum Kernel Maximum Mean Discrepancy

The Maximum Mean Discrepancy (MMD) provides a principled framework for comparing probability distributions by embedding them into a Reproducing Kernel Hilbert Space (RKHS). This approach presents a powerful, alignment-free metric for comparing generated sequences against reference distributions when combined with the Spectrum kernel for protein sequences.

**Theoretical Foundation.** The Maximum Mean Discrepancy (MMD) measures how different two probability distributions are by embedding them into a rich function space called a Reproducing Kernel Hilbert Space (RKHS). Intuitively, MMD finds the function that best distinguishes samples from the two distributions and measures how different the average function values are (Gretton et al., 2012).

Given samples $X = \{x_1, ..., x_n\}$ from a reference distribution and $Y = \{y_1, ..., y_m\}$ from a generated distribution, the squared MMD decomposes into three intuitive terms:

$$\texttt{MMD}^2 = \underbrace{\mathbb{E}[k(x, x')]}_{\text{similarity within } X} + \underbrace{\mathbb{E}[k(y, y')]}_{\text{similarity within } Y} - 2 \underbrace{\mathbb{E}[k(x, y)]}_{\text{similarity between } X \text{ and } Y}, \tag{25}$$

where $k(\cdot, \cdot)$ is a kernel function measuring sequence similarity. When sequences within each set are similar to each other (high first two terms) but the two sets differ from each other (low cross-term), the MMD is large, indicating distributional mismatch. Conversely, when generated sequences are indistinguishable from reference sequences, MMD approaches zero.

A key theoretical guarantee is that when using a *characteristic* kernel, MMD equals zero if and only if the two distributions are identical. This ensures that any systematic difference between generated and reference sequences will be detected, making MMD a principled choice for evaluating generative models.

In practice, the empirical MMD can be computed using either a biased (V-statistic) or unbiased (U-statistic) estimator, differing in whether diagonal terms of the kernel matrices are included. We use the biased estimator, which guarantees non-negative $\texttt{MMD}^2$ values, which matters when taking the square root, and exhibits lower variance than the unbiased alternative (Gretton et al., 2012).

**The Spectrum kernel**, introduced by Leslie et al. (2002), represents sequences as vectors of $k$-mer counts and computes similarity via their inner product. For protein sequences, this provides an alignment-free measure of compositional similarity.

Given a sequence $x$, we define the feature map $\Phi_k : \mathcal{X} \to \mathbb{R}^{|\mathcal{A}|^k}$ where $\mathcal{A}$ denotes the amino acid alphabet (typically 20 standard residues). Each coordinate of $\Phi_k(x)$ counts the occurrences of the corresponding $k$-mer:

$$\Phi_k(x)_a = |\{i : x[i : i + k] = a\}|. \tag{26}$$

This counts how many times the $k$-mer $a$ appears as a contiguous substring in sequence $x$, where $x[i : i + k]$ denotes the substring starting at position $i$ with length $k$.

The Spectrum kernel is then the inner product of these feature vectors:

$$k(x, y) = \langle \Phi_k(x), \Phi_k(y) \rangle = \sum_{a \in \mathcal{A}^k} \Phi_k(x)_a \cdot \Phi_k(y)_a. \tag{27}$$

This kernel has several attractive properties for protein sequence analysis:

1. It is efficient to compute using sparse representations, avoiding the need to explicitly enumerate all $|\mathcal{A}|^k$ possible $k$-mers.
2. It makes no assumptions about the data distribution, unlike learned embeddings.
3. It captures local sequence composition.
4. Its simplicity ensures robustness when evaluating artificial sequences (Kucera et al., 2022).

### D.6 Reference Profile Metrics

Two further metrics compare a generated set against the amino acid profile of the holdout set. Generated sequences are first aligned to the holdout alignment, giving for each column $i$ a reference distribution $p_i$ over the 20 amino acids and the gap, and a corresponding distribution $q_i$ for the generated set.

**Profile log-likelihood.** The log-likelihood of a sequence under the holdout profile,

$$\mathrm{LL}(x) = \sum_i \log p_i(x_i), \tag{28}$$

summed over aligned positions and averaged over generated sequences. Higher values indicate sequences that are more typical of the natural family.

**Entropy delta.** The difference in Shannon entropy between the generated and reference columns,

$$\Delta H_i = H(q_i) - H(p_i), \quad H(p) = -\sum_a p(a) \log p(a), \tag{29}$$

averaged over positions. A positive value means the generated set is more variable at that position than the natural family, and values closer to zero indicate a better match to the diversity of the holdout set.

## E Protein Types and Datasets

We use the following seed proteins to construct homolog datasets via iterative profile search. They were selected from systems available in our laboratory, so that experimental validation would be possible. They span antibody fragments, enzymes, and a growth factor, with mean sequence lengths from 109 to 363 residues and homolog sets ranging from roughly 1.4k to 24k sequences (Table 1).

### Anti-SARS-CoV-2 VHH (Ty1)

Ty1 is an alpaca-derived single-domain antibody (nanobody) that targets the receptor-binding domain (RBD) of the SARS-CoV-2 spike protein, directly preventing ACE2 engagement and neutralizing viral infection (Hanke et al., 2020). The 12.8 kDa nanobody binds an epitope accessible in both the "up" and "down" RBD conformations, sterically hindering host receptor binding. Cryo-electron microscopy structures reveal that CDR1 and CDR3 loops mediate the primary contacts with the spike protein.

### Serine-Pyruvate Aminotransferase (SPAT)

Serine-pyruvate aminotransferase (EC 2.6.1.51) is a pyridoxal phosphate-dependent enzyme that catalyzes the reversible transamination between L-serine and pyruvate to produce 3-hydroxypyruvate and L-alanine (BRENDA Enzyme Database, 2025). In humans, the enzyme localizes to peroxisomes where it participates in glyoxylate detoxification; functional deficiency causes primary hyperoxaluria type 1, characterized by calcium oxalate accumulation. The enzyme plays a key role in gluconeogenesis from serine in the liver.

### Haloalkane Dehalogenase DhaA

DhaA is a haloalkane dehalogenase from *Rhodococcus rhodochrous* that catalyzes the hydrolytic cleavage of carbon-halogen bonds in halogenated compounds (Koudelakova et al., 2013). The enzyme employs a catalytic pentad (Asp106, His272, Glu130, Asn41, Trp107) to perform nucleophilic substitution, forming a covalent alkyl-enzyme intermediate that is subsequently hydrolyzed. DhaA and its engineered variants are of biotechnological interest for bioremediation of industrial pollutants such as 1,2,3-trichloropropane.

### Anti-HER2 scFv

Single-chain variable fragments (scFvs) targeting HER2 consist of the variable heavy (VH) and light (VL) chains of an anti-HER2 antibody connected by a flexible glycine-serine linker (Koçer et al., 2021). HER2 (human epidermal growth factor receptor 2) is overexpressed in 20-25% of breast cancers, making it a critical target for molecular diagnostics and therapy. Anti-HER2 scFvs are used in antibody-drug conjugates and as targeting moieties for nanoparticle delivery systems.

### Chicken FGF2

Fibroblast growth factor 2 (FGF2) from *Gallus gallus* is a multifunctional growth factor involved in cell proliferation, differentiation, angiogenesis, and tissue repair (UniProt Consortium, 2025). The protein signals through FGF receptors 1-4 and is essential for normal embryonic development. In chickens, three FGF2 isoforms (18.5, 20.0, and 21.5 kDa) are produced by alternative translation initiation during embryogenesis.

### Anti-EphA2 VHH

Table 1: Summary statistics of homolog datasets used for training and evaluation. Query and target coverage values represent the fraction of the query and target sequences aligned in each homolog pair.

| Dataset | Number of Homologs | Sequence Length | Query Coverage | Target Coverage |
|---|---|---|---|---|
| Anti-SARS-CoV-2 VHH | 3335 | 109.2 ± 4.3 | 0.88 ± 0.01 | 1.00 ± 0.01 |
| Anti-EphA2 VHH | 24304 | 118.2 ± 3.7 | 1.00 ± 0.00 | 1.00 ± 0.00 |
| Serine-Pyruvate Aminotransferase | 6565 | 363.0 ± 16.7 | 0.93 ± 0.03 | 0.91 ± 0.10 |
| Chicken FGF2 | 1435 | 141.6 ± 11.4 | 0.88 ± 0.06 | 0.71 ± 0.19 |
| Haloalkane Dehalogenase DhaA | 4697 | 282.2 ± 11.5 | 0.96 ± 0.03 | 0.91 ± 0.10 |
| Anti-HER2 scFv | 10979 | 246.4 ± 4.8 | 0.99 ± 0.01 | 1.00 ± 0.00 |

Anti-EphA2 nanobodies are single-domain antibodies targeting ephrin type-A receptor 2, a receptor tyrosine kinase overexpressed in breast, prostate, lung, and bladder cancers (Roovers et al., 2011). Their small size (approximately 15 kDa), high stability, and superior tissue penetration make them well-suited for targeting solid tumors. These nanobodies have been developed for immunoliposomal drug delivery and as targeting agents for antibody-directed nanotherapeutics.

**Dataset Statistics** The number of homologs, sequence length distribution, and other relevant statistics for each seed protein dataset are detailed in Table 1.

## F Experimental Methods

### F.1 Binding kinetics

For the anti-SARS-CoV-2 VHH, binding kinetics were measured by SPR on the Carterra LSAXT using a SAHC30M chip. The chip was first conditioned with 1 minute injections of 10 mM glycine pH 2.0, 1X HBSTE (10 mM HEPES pH 7.4, 150 mM NaCl, 3 mM EDTA, 0.05% Tween-20) running buffer, and 20 mM NaOH. Next, an anti-his capture lawn was prepared by injecting a prepared solution of 75 ng/µL biotinylated THE His Tag Antibody (GenScript, A00613) over the sensor chip for 10 minutes. In general, a capture level of approximately 2000 RU for the anti-his lawn was achieved.

SARS-CoV-2 spike recombinant binding domain (R&D Systems, 10542CV100) and SARS-CoV-2 BA.2 spike recombinant binding domain (R&D Systems, 11148CV100) were used as antigens. Antigen dilutions were prepared in 1X HBSTE + 1% PEG3350. When antigens contained high refractive index stabilizers in the commercial product, such as glycerol or trehalose, antigens were buffer exchanged using Vivaspin columns (Cytiva, 28932235 or 28932236) per the manufacturer's instructions. Antigens after buffer exchange were quantified by bicinchoninic acid (BCA) assay (Merck, 71285-3) against a BSA standard curve. Antigen concentrations were optimized for each antigen (starting concentration from 250 nM to 1 µM, serial dilution factor of 1:3 to 1:5), for a total of 6 antigen concentrations for each run.

Capture kinetics experiments were performed using a running buffer of 1X HBSTE + 1% PEG3350 with the chip at 25°C. Variant libraries, prepared in a dilution of anywhere from 1:20 to 1:200 in 1X HBSTE depending on the ideal capture level for kinetics (generally 150 to 200 RU, to achieve an Rmax of approximately 50 to 100 RU) in deep 384-well plates, were captured on the chip for 5 minutes. 5 injections of 1X HBSTE + 1% PEG3350 were first applied to achieve a stable baseline, followed by injections of the antigens in ascending concentrations, using a 5 minute association and 10 minute dissociation. After all concentrations were injected, the chip surface was regenerated by 3 x 30 second pulses with 10 mM glycine pH 2.0.

Raw kinetic data were analyzed using the Carterra Kinetics software 1.9.1.4215 using a non-regenerative kinetics approach. Ligand (variant) injections were y-aligned and capture levels were determined by setting a report point after ligand capture. Curves were double-referenced, first by subtracting a negative control spot where no binding was expected to account for non-specific background interactions, followed by blanking with an injection of running buffer alone. A 1:1 Langmuir fit was applied to calculate the $k_d$, $k_a$, and $K_D$ of each variant.

### F.2 Thermostability

To measure melting temperature, the Prometheus NT.48 instrument (NanoTemper Technologies) was used. Standard capillaries (PR-C002, NanoTemper Technologies) were filled with approximately 10 µL of each purified protein sample and loaded into the sample drawer 48 at a time. Temperature was increased from 20°C to 80°C or 90°C by 1°C $\text{min}^{-1}$. Melting temperatures were called manually based on the peaks observed from the first derivative of the ratio of the absorbance at 350 nm over the absorbance at 330 nm.

### F.3 Haloalkane dehalogenase activity

The enzymatic activity was measured using a fluorescence-based spectrophotometric assay as described by Dockalova et al. (2020). The assay monitors the increase in fluorescence intensity resulting from the cleavage of the halide ion from 4-bromomethyl-6,7-dimethoxycoumarin (COU-Br; Sigma-Aldrich, Cat. No. 301450) to yield the corresponding fluorescent product. COU-Br was dissolved in DMSO to prepare a 1.25 mM working stock solution. Purified proteins were normalized to a final concentration of 0.16 µM and diluted in an enzyme mix containing 50 mM phosphate buffer (pH 7.4). For each reaction, 2 µL of 1.25 mM COU-Br stock solution was added to wells of a 384-well black microtiter plate (Corning, Cat. No. 3575). The reaction was initiated by adding 18 µL of the enzyme mix to each well, resulting in a final COU-Br concentration of 0.125 mM and a total reaction volume of 20 µL. Fluorescence intensity was monitored continuously for 20 minutes at 30°C using a Spark Multimode Microplate Reader with excitation and emission wavelengths set to 345 nm and 437 nm, respectively (bandwidth 10 nm for both). Measurements were taken from the top of the plate. All assays were performed with two technical replicates and two biological replicates. Initial reaction rates were determined from the linear portion of the fluorescence curve within the first 200 seconds.

## G Deterministic Benchmark

This benchmark isolates the core capability behind EvoFlows: whether an edit-flow model can learn a sequence-to-sequence mapping, recovering the specific edits that connect a source to its target, rather than generating a sequence from noise as in standard flow matching. We construct a synthetic dataset with deterministic mutation patterns to evaluate this in a controlled setting. From a natural protein sequence $z_0$, we generate $z_1$ through position-dependent rules:

- If $z_0[i] = \text{A}$, include $\text{Sub}(i + 5, \text{H})$.
- If $z_0[i] = \text{C}$, include $\text{Ins}(i - 2, \text{S})$.

Table 2: *Mutation classification on the deterministic benchmark.* Precision, recall, and F1-score for each class (substitution, insertion, deletion, and no-op), along with ground-truth class prevalence. Values are reported with bootstrap confidence intervals. Performance is highest for no-ops, while insertion, substitution, and deletion remain well separated, indicating reliable discrimination between mutation types in the deterministic setting.

| **Mutation Type** | **Precision** | **Recall** | **F1-Score** | **Prevalence** |
|---|---|---|---|---|
| No-op | $0.982 \pm 0.001$ | $0.983 \pm 0.000$ | $0.982 \pm 0.000$ | 0.888 |
| Insertion | $0.820 \pm 0.005$ | $0.796 \pm 0.006$ | $0.808 \pm 0.005$ | 0.047 |
| Substitution | $0.804 \pm 0.005$ | $0.843 \pm 0.007$ | $0.823 \pm 0.005$ | 0.043 |
| Deletion | $0.910 \pm 0.006$ | $0.850 \pm 0.007$ | $0.879 \pm 0.006$ | 0.022 |

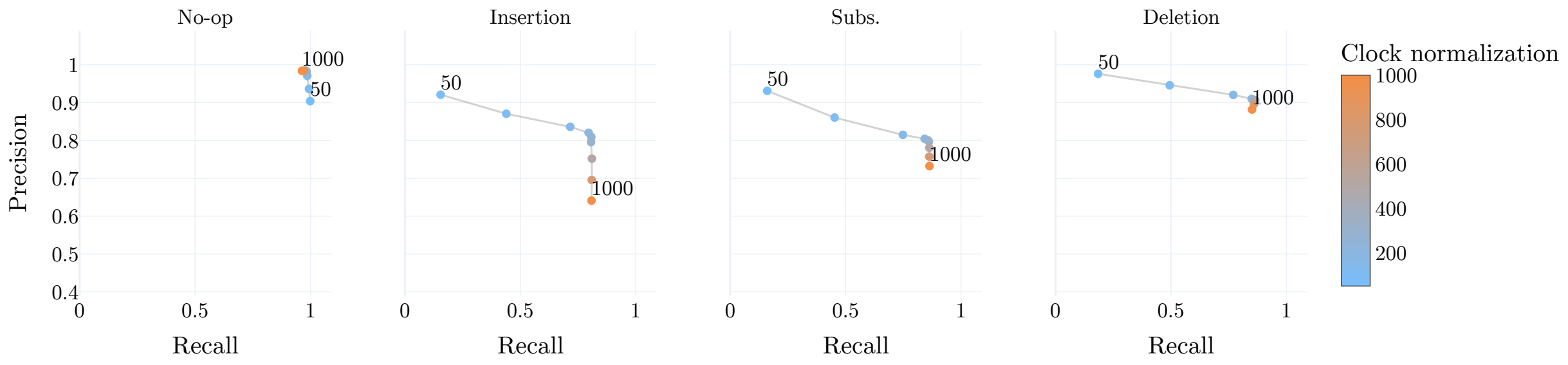

Figure 4: *Precision-recall trade-offs under clock normalization.* Precision-recall curves for each mutation type and no-ops on the deterministic benchmark, which shows how clock normalization controls the trade-off between recall and precision. While no-op predictions remain stable, insertion, substitution, and deletion exhibit systematic precision-recall shifts as the clock normalization varies, highlighting its role in calibrating mutation selection behavior.

- If $z_0[i] = \mathrm{G}$ and $z_0[j] = \mathrm{L}$ and $z_0[k] = \mathrm{K}$ for some $j < i < k$, include Del($i$).

Mutations are applied in order: insertions (increasing index), deletions, then substitutions. This yields a unique $z_1$ for each $z_0$, enabling precise assessment of mutation type, position, and amino acid identity, unlike natural MSAs where alignments may admit multiple explanations. Results across sequence lengths and clock normalization factors are shown in Figure 4, Figure 5, and Table 2. The clock normalization applies a length-normalized rate scaling (Section 3.3) and is calibrated automatically rather than set by hand, so Table 2 reports the operating point that this calibration produces, while Figure 4 sweeps the scaling to show how it trades precision against recall for each edit type.

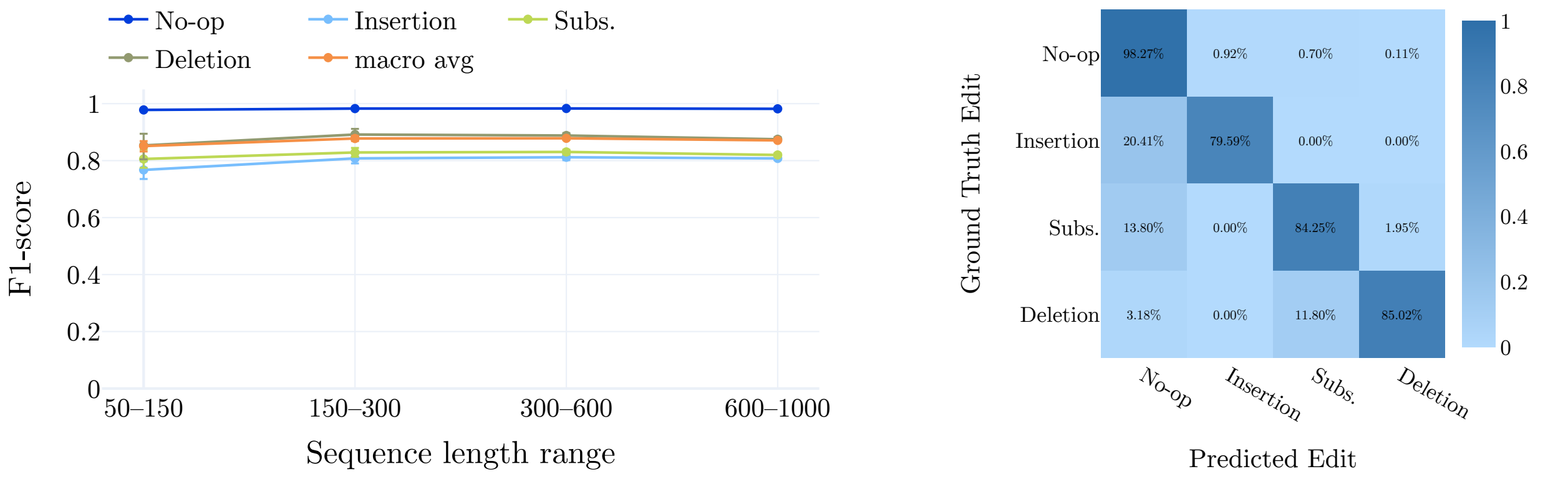

Figure 5: *Performance on the deterministic benchmark. (Left)* F1-score for each mutation type as a function of sequence length. Performance remains stable across sequence lengths, indicating that mutation prediction accuracy does not degrade for longer sequences. *(Right)* Confusion matrix showing the distribution of predicted mutation types versus ground-truth mutations (no-op, insertion, substitution, deletion). The model correctly identifies most operations, with remaining errors primarily corresponding to confusions with the no-op class.

## H Distribution of Generated Variants

We summarize the generated variant sets for the two proteins used in our *in vitro* validation (Section 4.3): the anti-SARS-CoV-2 VHH binder and the DhaA enzyme. Edit positions are reported in template coordinates, so the axis is the length of the template and an insertion is attributed to the residue it follows. Sequence lengths are ungapped.

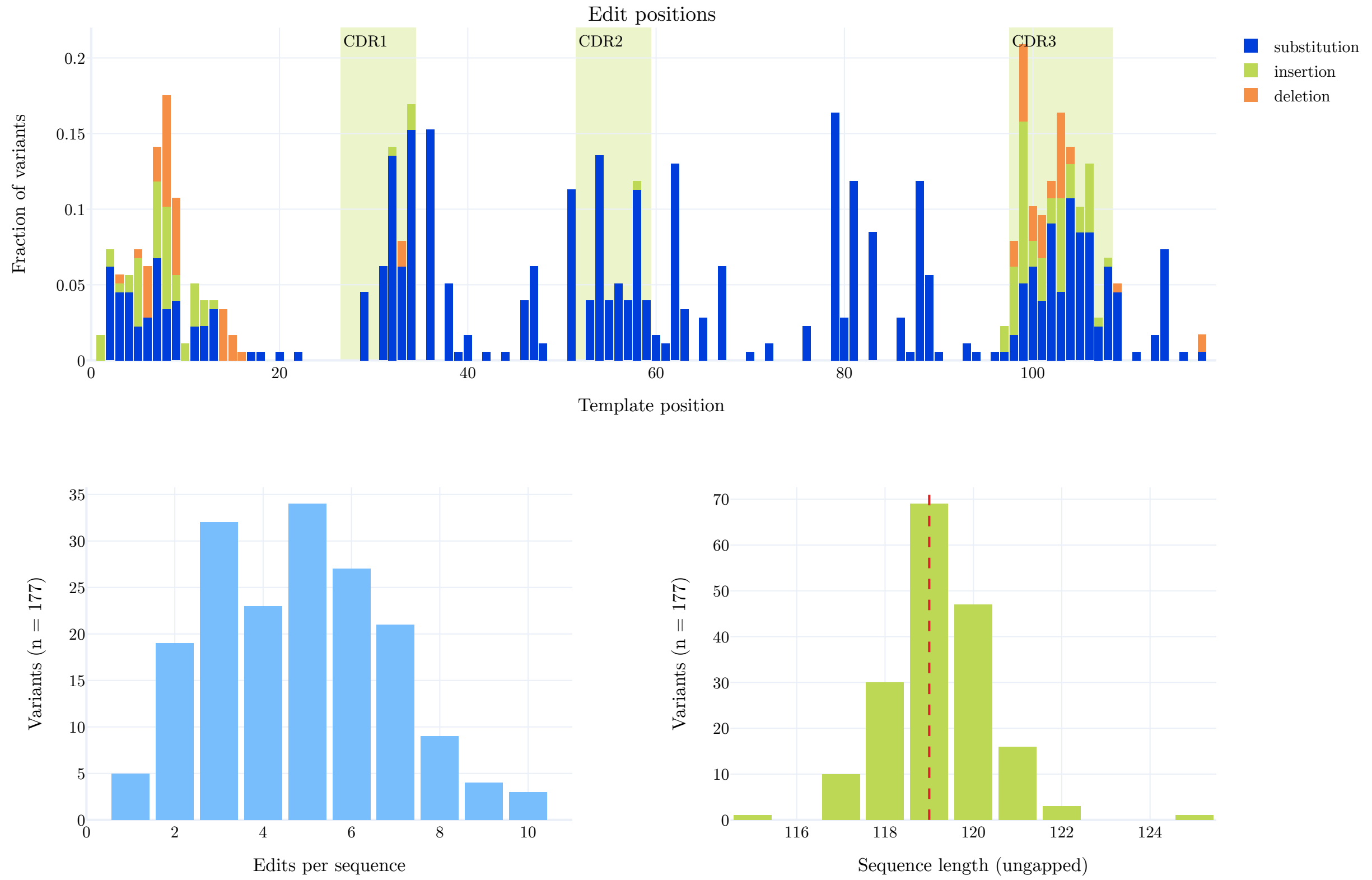

Figure 6: *Binder: what EvoFlows generated. (Top)* Fraction of variants carrying each edit type at each template position, stacked by type, with the CDRs shaded. Positions are 1-based residue numbers and an insertion is drawn at the residue it follows. *(Bottom)* Edits per sequence, and ungapped sequence length with the template marked by the dashed line.

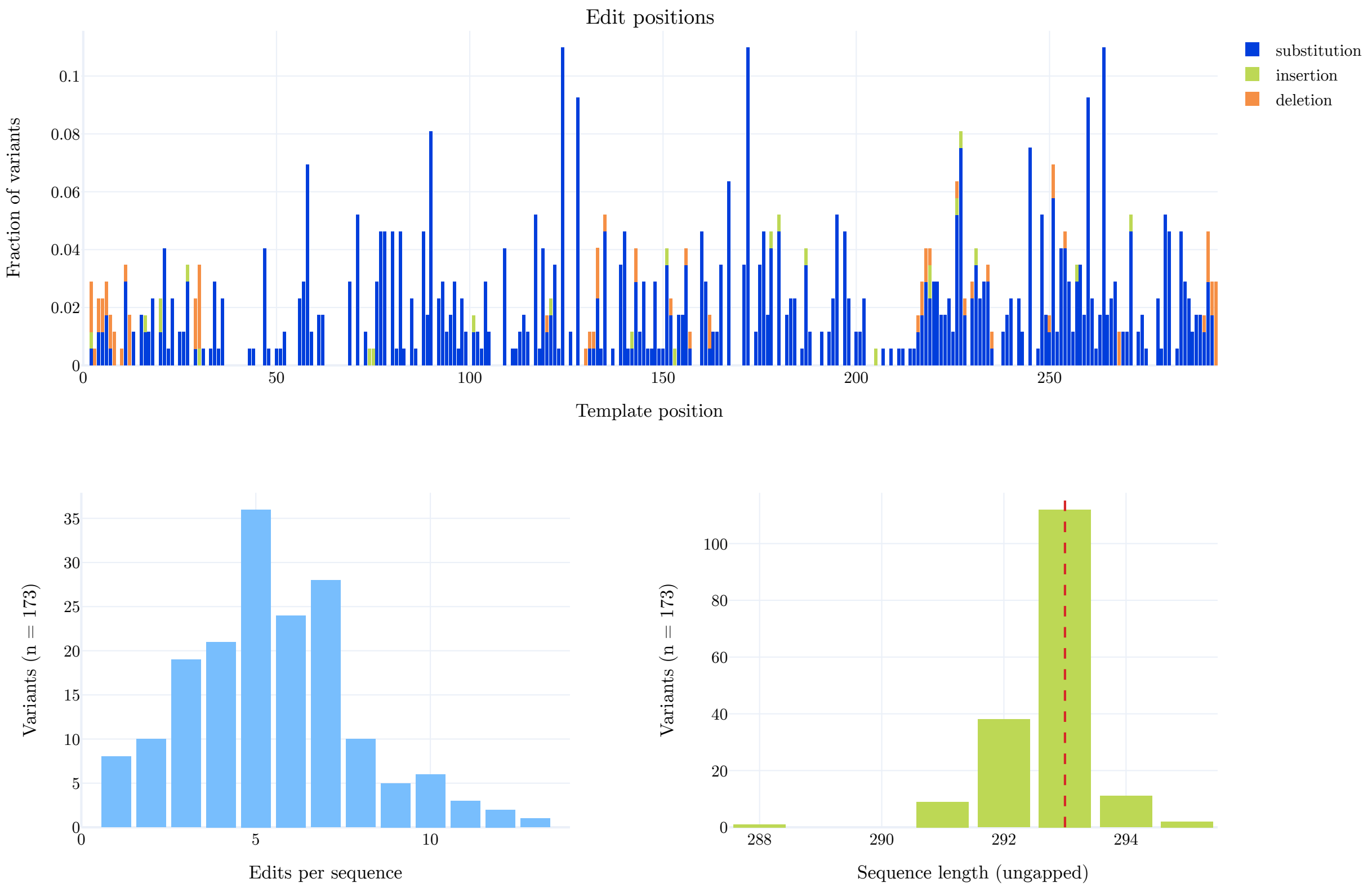

Figure 7: *Enzyme (DhaA): what EvoFlows generated.* Panels as in Figure 6; DhaA has no CDRs to shade.

## I Supplementary Figures

Table 3: Numerical values underlying Figure 2, aggregated across the six homolog datasets. The standard deviation is taken across datasets, which differ in family and sequence length, so it reflects dataset heterogeneity rather than uncertainty in the comparison between methods; per-dataset values are given in Figure 8. Random pairing draws both sequences from the natural homolog distribution and acts as an upper reference rather than a competing method. †Evotune and EvoDiff-MSA introduce less than half as many edits as the remaining methods, so their distributional scores are not directly comparable: a method that barely changes the template trivially remains close to the natural distribution. The two Evotune columns isolate this effect, since they are the same model: forcing it to a comparable edit count degrades every distributional metric. EvoFlows reaches a similar edit count while degrading far less, and among the methods with comparable edit budgets (EvoFlows, Evotune with forced substitutions, and random mutations) it leads on every metric.

| **Metric** | **Random pairing** | **EvoFlow (ours)** | **EvoDiff-MSA†** | **Evotune†** | **Evotune (forced)** | **Random** |
|---|---|---|---|---|---|---|
| Avg Levenshtein to x0 | 116 ± 87.1 | 31.7 ± 19.80 | 13.2 ± 8.91 | 13.3 ± 14.67 | 34.0 ± 23.99 | 31.1 ± 20.16 |
| Avg pairwise Levenshtein | 116 ± 87.1 | 117 ± 84.7 | 119 ± 90.8 | 118 ± 87.7 | 128 ± 86.4 | 139 ± 87.6 |
| Covariance | 0.99 ± 0.009 | 0.97 ± 0.026 | 0.98 ± 0.008 | 0.99 ± 0.008 | 0.95 ± 0.043 | 0.90 ± 0.097 |
| ESM2 PLL | −1.3 ± 0.15 | −1.3 ± 0.18 | −1.4 ± 0.27 | −1.3 ± 0.17 | −1.5 ± 0.17 | −2.3 ± 0.27 |
| Entropy delta | 0.03 ± 0.041 | 0.03 ± 0.054 | 0.05 ± 0.041 | 0.04 ± 0.042 | 0.13 ± 0.090 | 0.24 ± 0.135 |
| JS divergence | 0.00 ± 0.001 | 0.00 ± 0.002 | 0.00 ± 0.001 | 0.00 ± 0.001 | 0.01 ± 0.005 | 0.02 ± 0.012 |
| KL divergence | 0.01 ± 0.004 | 0.02 ± 0.007 | 0.01 ± 0.004 | 0.01 ± 0.004 | 0.03 ± 0.017 | 0.07 ± 0.040 |
| MIP | 0.92 ± 0.068 | 0.88 ± 0.065 | 0.89 ± 0.062 | 0.92 ± 0.070 | 0.80 ± 0.101 | 0.66 ± 0.233 |
| MMD | 0.64 ± 0.298 | 1.1 ± 0.23 | 0.76 ± 0.308 | 0.68 ± 0.309 | 1.7 ± 0.52 | 2.3 ± 0.82 |
| Profile log-likelihood | −328 ± 241.8 | −325 ± 236.0 | −345 ± 263.9 | −336 ± 246.3 | −377 ± 241.8 | −500 ± 321.9 |

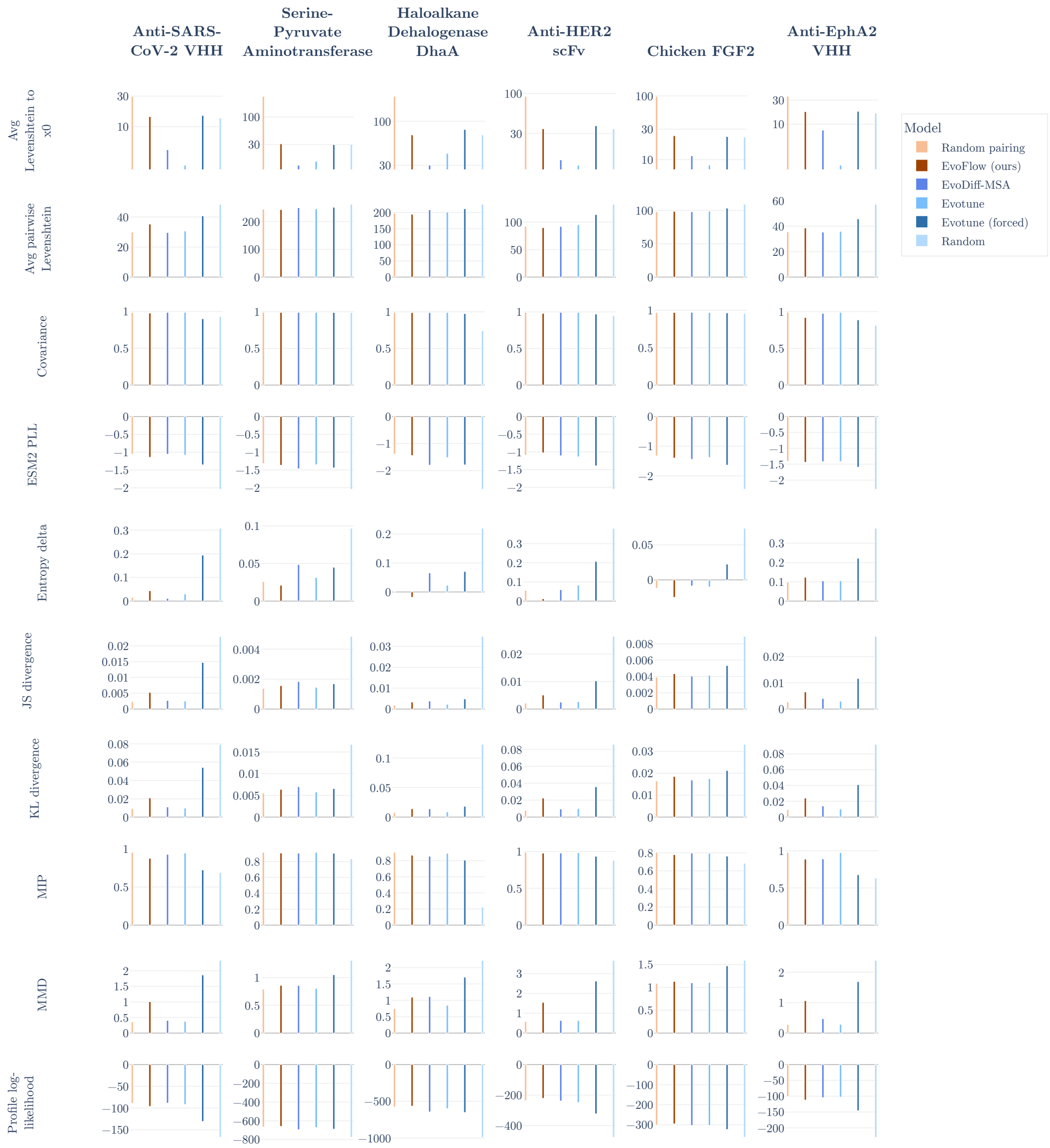

Figure 8: *Per-dataset comparison of EvoFlows and baseline methods.* Comparison across multiple datasets (columns) and evaluation metrics (rows). The random transport model produces the most different sequences from the starting sequences as a function of the data. EvoFlows generates variants that remain close to the holdout distribution. Random baseline performs worst across all metrics. Evotuning without forced mutations achieves competitive scores but introduces almost no sequence mutations, as reflected in Levenshtein distances to x0, while forcing mutations significantly degrades performance across metrics.

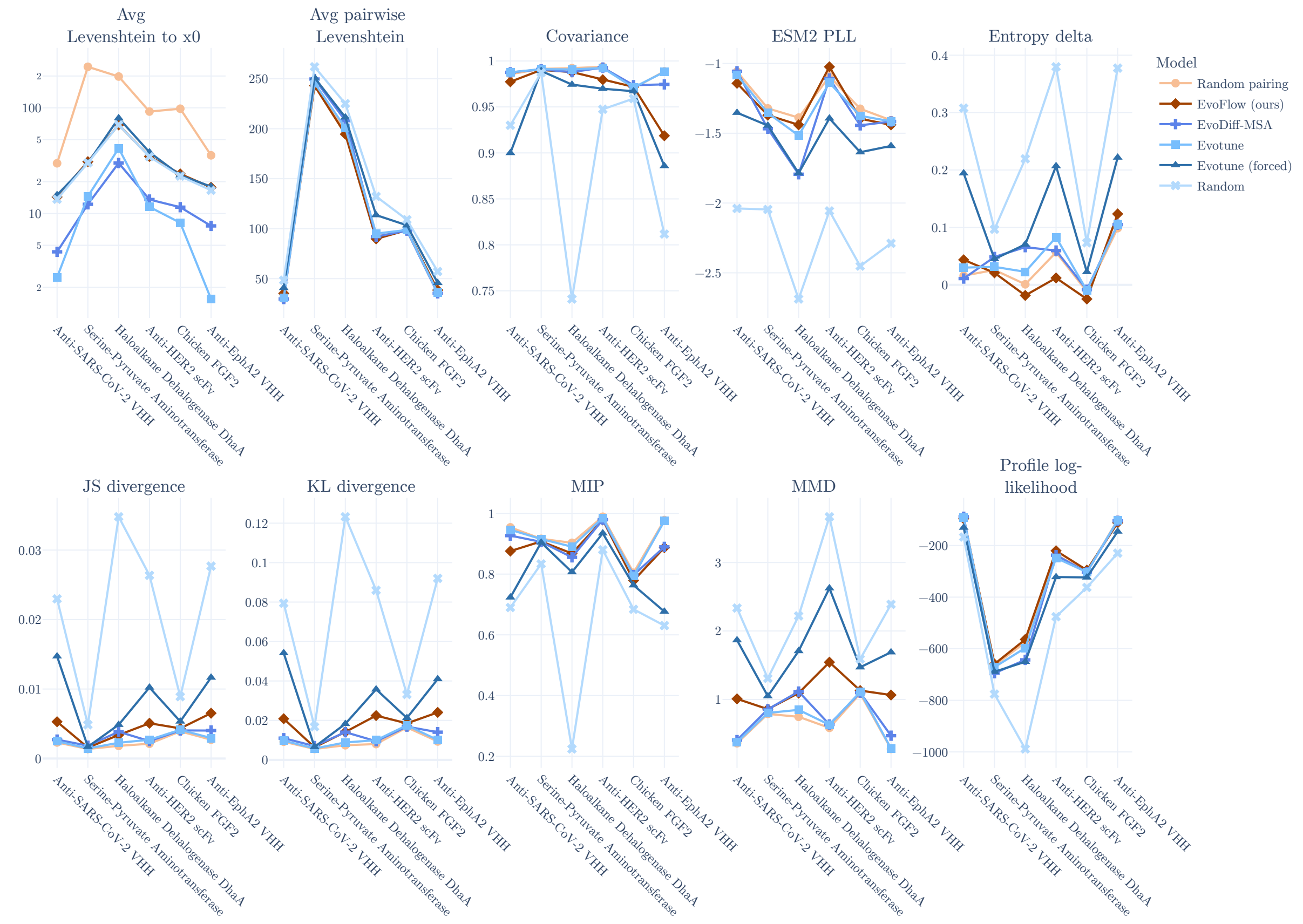

Figure 9: *Cross-dataset metric trends.* Each subplot shows one evaluation metric, with lines connecting per-dataset values for each method: Random pairing, EvoFlows (ours), Evotuning, Evotuning with forced mutations, and random baseline.

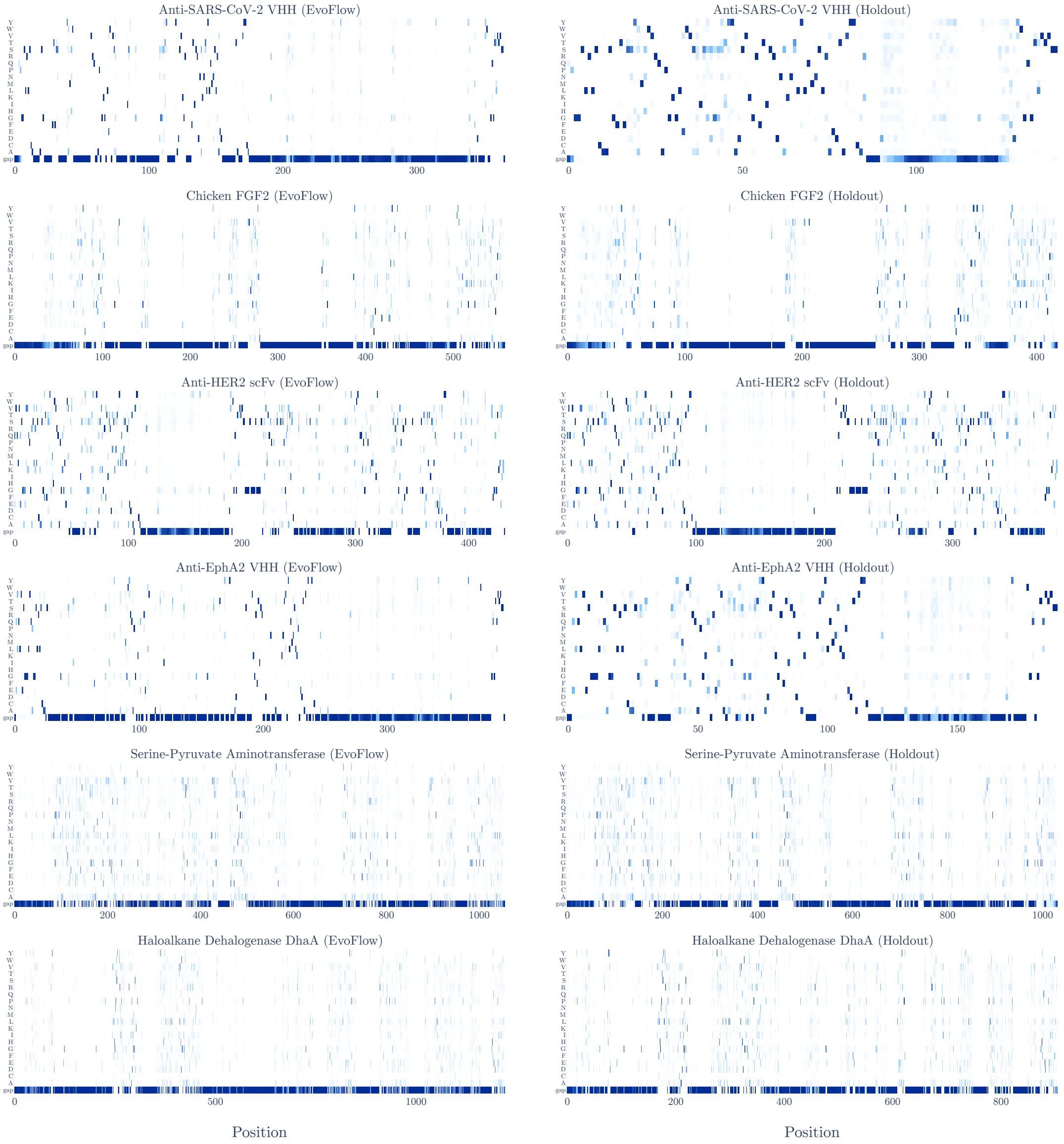

Figure 10: *Per-position amino acid frequency heatmaps for all seed protein families.* Each row shows one dataset, with EvoFlows-generated sequences (left) and holdout sequences (right). Color intensity indicates per-position amino acid frequency after alignment. Conserved positions appear as points of high intensity, while variable regions show more diffuse patterns. The similar frequency profiles between generated and holdout sequences indicate that EvoFlows preserves the positional amino acid distribution characteristic of each homolog family.